\documentclass[11pt,a4paper]{article}
\usepackage[hyperref]{emnlp-ijcnlp-2019}
\usepackage{float}
\usepackage{times}
\usepackage{latexsym}
\usepackage{url}
\usepackage{bm}
\usepackage{graphicx}
\usepackage{subfig}

\newlength\lengtha 
\newlength\lengthb 
\newlength\lengthc 
\newlength\lengthd 

\usepackage{multirow}
\usepackage{amsmath}
\usepackage{capt-of}
\usepackage{tabularx}
\usepackage{amssymb}
\usepackage{amsfonts}
\usepackage{booktabs}
\usepackage{scalerel}
\usepackage[inline]{enumitem}
\usepackage{listings}
\usepackage{varwidth}
\usepackage[export]{adjustbox}
\usepackage{tikz}
\usetikzlibrary{tikzmark}
\usepackage{todonotes}
\usepackage{cleveref}

\usepackage{stmaryrd}
\usepackage{bbm}
\usepackage{graphicx}

\newcommand{\tabincell}[2]{\begin{tabular}{@{}#1@{}}#2\end{tabular}}

\usepackage{algorithm}
\usepackage[noend]{algpseudocode}

\definecolor{deepblue}{rgb}{0,0,0.5}
\definecolor{officeblue}{RGB}{0,102,204}
\definecolor{deepred}{rgb}{0.6,0,0}
\definecolor{deepgreen}{rgb}{0,0.5,0}
\definecolor{mybrickred}{RGB}{182,50,28}

\definecolor{fillcolor}{RGB}{216,217,252}

\algnewcommand\algorithmicrequireb{{\hspace{0.95cm}}}
\algnewcommand\INPTDESCB{\item[\algorithmicrequireb]}

\algnewcommand\algorithmicfuncdesc{\textbf{Function:}}
\algnewcommand\FUNCDESC{\item[\algorithmicfuncdesc]}
\algnewcommand\algorithmicfuncdescb{{\hspace{0.86cm}}}
\algnewcommand\FUNCDESCB{\item[\algorithmicfuncdescb]}
\algnewcommand{\algorithmicgoto}{\textbf{goto}}
\algnewcommand{\Goto}[1]{\algorithmicgoto~\ref{#1}}

\usepackage{amsmath,amsfonts,bm}

\def\eqref#1{equation~\ref{#1}}

\def\1{\bm{1}}

\DeclareMathAlphabet{\mathsfit}{\encodingdefault}{\sfdefault}{m}{sl}
\SetMathAlphabet{\mathsfit}{bold}{\encodingdefault}{\sfdefault}{bx}{n}

\aclfinalcopy 

\title{Visualizing and Understanding the Effectiveness of BERT}

\author{Yaru Hao$^\dag$\thanks{\ \  Contribution during internship at Microsoft Research.},~~Li Dong$^\ddag$,~~Furu Wei$^\ddag$,~~Ke Xu$^{\dag}$\\
$^\dag$Beihang University \\
$^\ddag$Microsoft Research \\
\texttt{\{haoyaru@,kexu@nlsde.\}buaa.edu.cn} \\
\texttt{\{lidong1,fuwei\}@microsoft.com} \\}

\date{}

\begin{document}
\maketitle
\begin{abstract}
Language model pre-training, such as BERT, has achieved remarkable results in many NLP tasks. However, it is unclear why the pre-training-then-fine-tuning paradigm can improve performance and generalization capability across different tasks. In this paper, we propose to visualize loss landscapes and optimization trajectories of fine-tuning BERT on specific datasets. First, we find that pre-training reaches a good initial point across downstream tasks, which leads to wider optima and easier optimization compared with training from scratch. We also demonstrate that the fine-tuning procedure is robust to overfitting, even though BERT is highly over-parameterized for downstream tasks. Second, the visualization results indicate that fine-tuning BERT tends to generalize better because of the flat and wide optima, and the consistency between the training loss surface and the generalization error surface. Third, the lower layers of BERT are more invariant during fine-tuning, which suggests that the layers that are close to input learn more transferable representations of language.
\end{abstract}

\section{Introduction}

Language model pre-training has achieved strong performance in many NLP tasks~\cite{elmo,ulmfit,gpt,bert,clozepretrain19,unilm}.
A neural encoder is trained on a large text corpus by using language modeling objectives. Then the pre-trained model either is used to extract vector representations for input, or is fine-tuned on the specific datasets.

Recent work~\cite{probing1,probing2,assessingBert,pipline} has shown that the pre-trained models can encode syntactic and semantic information of language.
However, it is unclear why pre-training is effective on downstream tasks in terms of both trainability and generalization capability.
In this work, we take BERT~\cite{bert} as an example to understand the effectiveness of pre-training.
We visualize the loss landscapes and the optimization procedure of fine-tuning on specific datasets in three ways.
First, we compute the one-dimensional (1D) loss curve, so that we can inspect the difference between fine-tuning BERT and training from scratch.
Second, we visualize the two-dimensional (2D) loss surface, which provides more information about loss landscapes than 1D curves.
Third, we project the high-dimensional optimization trajectory of fine-tuning to the obtained 2D loss surface, which demonstrate the learning properties in an intuitive way.

The main findings are summarized as follows.
First, visualization results indicate that BERT pre-training reaches a good initial point across downstream tasks, which leads to wider optima on the 2D loss landscape compared with random initialization.
Moreover, the visualization of optimization trajectories shows that pre-training results in easier optimization and faster convergence. We also demonstrate that the fine-tuning procedure is robust to overfitting.
Second, loss landscapes of fine-tuning partially explain the good generalization capability of BERT. Specifically, pre-training obtains more flat and wider optima, which indicates the pre-trained model tends to generalize better on unseen data~\cite{widevalley,visualloss,izmailov2018averaging}.
Additionally, we find that the training loss surface correlates well with the generalization error.
Third, we demonstrate that the lower (i.e., close to input) layers of BERT are more invariant across tasks than the higher layers, which suggests that the lower layers learn transferable representations of language. We verify the point by visualizing the loss landscape with respect to different groups of layers.

\section{Background: BERT}

We use BERT (Bidirectional Encoder Representations from Transformers;~\citealt{bert}) as an example of pre-trained language models in our experiments.
BERT is pre-trained on a large corpus by using the masked language modeling and next-sentence prediction objectives.
Then we can add task-specific layers to the BERT model, and fine-tune all the parameters according to the downstream tasks.

BERT employs a Transformer~\cite{transformer} network to encode contextual information, which contains multi-layer self-attention blocks.
Given the embeddings $\{\textbf{x}_i\}_{i=1}^{|x|}$ of input text, we concatenate them into $\mathbf{H}^0 = [\mathbf{x}_1, \cdots, \mathbf{x}_{|x|}]$. Then, an $L$-layer Transformer encodes the input: $\mathbf{H}^l = \mathrm{Transformer\_block}_{l}(\mathbf{H}^{l-1})$, where $l = 1, \cdots, L$, and $\mathbf{H}^L = [\mathbf{h}_1^L, \cdots, \mathbf{h}_{|x|}^L]$.
We use the hidden vector $\mathbf{h}_i^L$ as the contextualized representation of the input token $x_i$.
For more implementation details, we refer readers to~\citet{transformer}.



\section{Methodology}

We employ three visualization methods to understand why fine-tuning the pre-trained BERT model can achieve better performance on downstream tasks compared with training from scratch.
We plot both one-dimensional and two-dimensional loss landscapes of BERT on the specific datasets. Besides, we project the optimization trajectories of the fine-tuning procedure to the loss surface.
The visualization algorithms can also be used for the models that are trained from random initialization, so that we can compare the difference between two learning paradigm.

\subsection{One-dimensional Loss Curve}
\label{sec:1d:loss}

Let $\bm{\theta}_0$ denote the initialized parameters.
For fine-tuning BERT, $\bm{\theta}_0$ represents the the pre-trained parameters.
For training from scratch, $\bm{\theta}_0$ represents the randomly initialized parameters.
After fine-tuning, the model parameters are updated to $\bm{\theta}_1$.
The one-dimensional (1D) loss curve aims to quantify the loss values along the optimization direction (i.e., from $\bm{\theta}_0$ to $\bm{\theta}_1$).

The loss curve is plotted by linear interpolation between $\bm{\theta}_0$ and $\bm{\theta}_1$~\cite{curve1d}.
The curve function $f(\alpha)$ is defined as:
\begin{equation}
f(\alpha) = \mathcal{J}(\bm{\theta}_0+\alpha \bm{\delta}_1) \label{eq:1d:loss}
\end{equation}
where $\alpha$ is a scalar parameter, $\bm{\delta}_1 = \bm{\theta}_1-\bm{\theta}_0$ is the optimization direction, and $\mathcal{J}(\bm{\theta})$ is the loss function under the model parameters $\bm{\theta}$.
In our experiments, we set the range of $\alpha$ to $[-4, 4]$ and sample $40$ points for each axis.
Note that we only consider the parameters of BERT in $\bm{\theta}_0$ and $\bm{\theta}_1$, so $\bm{\delta}_1$ only indicates the updates of the original BERT parameters. The effect of the added task-specific layers is eliminated by keeping them fixed to the learned values.

\subsection{Two-dimensional Loss Surface}
\label{subsc:loss_surface}

The one-dimensional loss curve can be extended to the two-dimensional (2D) loss surface~\cite{visualloss}.
Similar as in Equation~(\ref{eq:1d:loss}), we need to define two directions ($\bm{\delta}_1$ and $\bm{\delta}_2$) as axes to plot the loss surface:
\begin{equation}
f(\alpha, \beta) = \mathcal{J}(\bm{\theta}_0+\alpha \bm{\delta}_1+\beta \bm{\delta}_2)
\label{eq:2d:loss}
\end{equation}
where $\alpha,\beta$ are scalar values, $\mathcal{J}(\cdot)$ is the loss function, and $\bm{\theta}_0$ represents the initialized parameters. Similar to Section~\ref{sec:1d:loss}, we are only interested in the parameter space of the BERT encoder, without taking into consideration task-specific layers.
One of the axes is the optimization direction $\bm{\delta}_1 = \bm{\theta}_1-\bm{\theta}_0$ on the target dataset, which is defined in the same way as in Equation~(\ref{eq:1d:loss}).
We compute the other axis direction via $\bm{\delta}_2 = \bm{\theta}_2-\bm{\theta}_0$, where $\bm{\theta}_2$ represents the fine-tuned parameters on another dataset. So the other axis is the optimization direction of fine-tuning on another dataset.
Even though the other dataset is randomly chosen, experimental results confirm that the optimization directions $\bm{\delta}_1, \bm{\delta}_2$ are divergent and orthogonal to each other because of the high-dimensional parameter space.



The direction vectors $\bm{\delta}_1$ and $\bm{\delta}_2$ are projected onto a two-dimensional plane. It is beneficial to ensure the scale equivalence of two axes for visualization purposes.
Similar to the filter normalization approach introduced in~\cite{visualloss}, we address this issue by normalizing two direction vectors to the same norm.
We re-scale $\bm{\delta}_2$ to $\frac{\left \| \bm{\delta}_1 \right \|}{\left \| \bm{\delta}_2 \right \|} \bm{\delta}_2$, where $\left \| \cdot \right \|$ denotes the Euclidean norm.
We set the range of both $\alpha$ and $\beta$ to $[-4, 4]$ and sample $40$ points for each axis.

\subsection{Optimization Trajectory}
\label{subsc:opt_traj}

Our goal is to project the optimization trajectory of the fine-tuning procedure onto the 2D loss surface obtained in Section~\ref{subsc:loss_surface}.
Let $\{(d_i^\alpha, d_i^\beta)\}_{i=1}^{T}$ denote the projected optimization trajectory, where $(d_i^\alpha, d_i^\beta)$ is a projected point in the loss surface, and $i=1,\cdots,T$ represents the $i$-th epoch of fine-tuning.

As shown in Equation~(\ref{eq:2d:loss}), we have known the optimization direction $\bm{\delta}_1 = \bm{\theta}_1-\bm{\theta}_0$ on the target dataset.
We can compute the deviation degrees between the optimization direction and the trajectory to visualize the projection results.
Let $\bm{\theta}^i$ denote the BERT parameters at the $i$-th epoch, and $\bm{\delta}^i=\bm{\theta}^i-\bm{\theta}_0$ denote the optimization direction at the $i$-th epoch.
The point $(d_i^\alpha, d_i^\beta)$ of the trajectory is computed via:
\begin{align}
v_{cos} &= \frac{ \bm{\delta}^i \times \bm{\delta}_1 }{\left \| \bm{\delta}^i \right \| \cdot \left \| \bm{\delta}_1 \right \|} \\
d_i^\alpha &= v_{cos} \frac{\left \| \bm{\delta}^i \right \|}{\left \| \bm{\delta}_1 \right \|} = \frac{ \bm{\delta}^i \times \bm{\delta}_1 }{\left \| \bm{\delta}_1 \right \|^2} \\
d_i^\beta &= \sqrt{(\frac{\left \| \bm{\delta}^i \right \|}{\left \| \bm{\delta}_1 \right \|})^2 - (d_i^\alpha)^2} 
\end{align}
where $\times$ denotes the cross product of two vectors, and $\left \| \cdot \right \|$ denotes the Euclidean norm.
To be specific, we first compute cosine similarity between $\bm{\delta}^i$ and $\bm{\delta}_1$, which indicates the angle between the current optimization direction and the final optimization direction.
Then we get the projection values $d_i^\alpha$ and $d_i^\beta$ by computing the deviation degrees between the optimization direction $\bm{\delta}^i$ and the axes.

\section{Experimental Setup}

We conduct experiments on four datasets: Multi-genre Natural Language Inference Corpus (MNLI;~\citealt{mnli2017}), Recognizing Textual Entailment (RTE;~\citealt{rte1,rte2,rte3,rte5}), Stanford Sentiment Treebank (SST-2;~\citealt{sst2013}), and Microsoft Research Paraphrase Corpus (MRPC;~\citealt{mrpc2005}).
We use the same data split as in~\cite{wang2018glue}.
The accuracy metric is used for evaluation.

We employ the pre-trained BERT-large model in our experiments.
The cased version of tokenizer is used.
We follow the settings and the hyper-parameters suggested in~\cite{bert}.
The Adam~\cite{adam} optimizer is used for fine-tuning.
The number of fine-tuning epochs is selected from $\{3,4,5\}$.
For RTE and MRPC, we set the batch size to $32$, and the learning rate to 1e-5. For MNLI and SST-2, the batch size is $64$, and the learning rate is 3e-5.

For the setting of training from scratch, we use the same network architecture as BERT, and randomly initialize the model parameters.
Most hyper-parameters are kept the same. The number of training epochs is larger than fine-tuning BERT, because training from scratch requires more epochs to converge. The number of epochs is set to $8$ for SST-2, and $16$ for the other datasets, which is validated on the development set.

\section{Pre-training Gets a Good Initial Point Across Downstream Tasks}


Fine-tuning BERT on the usually performs significantly better than training the same network with random initialization, especially when the data size is small. Results indicate that language model pre-training objectives learn good initialization for downstream tasks.
In this section, we inspect the benefits of using BERT as the initial point from three aspects.

\subsection{Pre-training Leads to Wider Optima}
\label{subsc:wider_optima}

\begin{figure*}[t]
\centering
\small
\begin{tabular}{l@{\hspace*{\lengthc}}c@{\hspace*{\lengthc}}c@{\hspace*{\lengthc}}c@{\hspace*{\lengthc}}c@{\hspace*{\lengthc}}}
&MNLI&RTE&SST-2&MRPC\\ 
\rotatebox[origin=c]{90}{Training from scratch}&
\raisebox{-0.5\height}{\includegraphics[width=0.5\columnwidth]{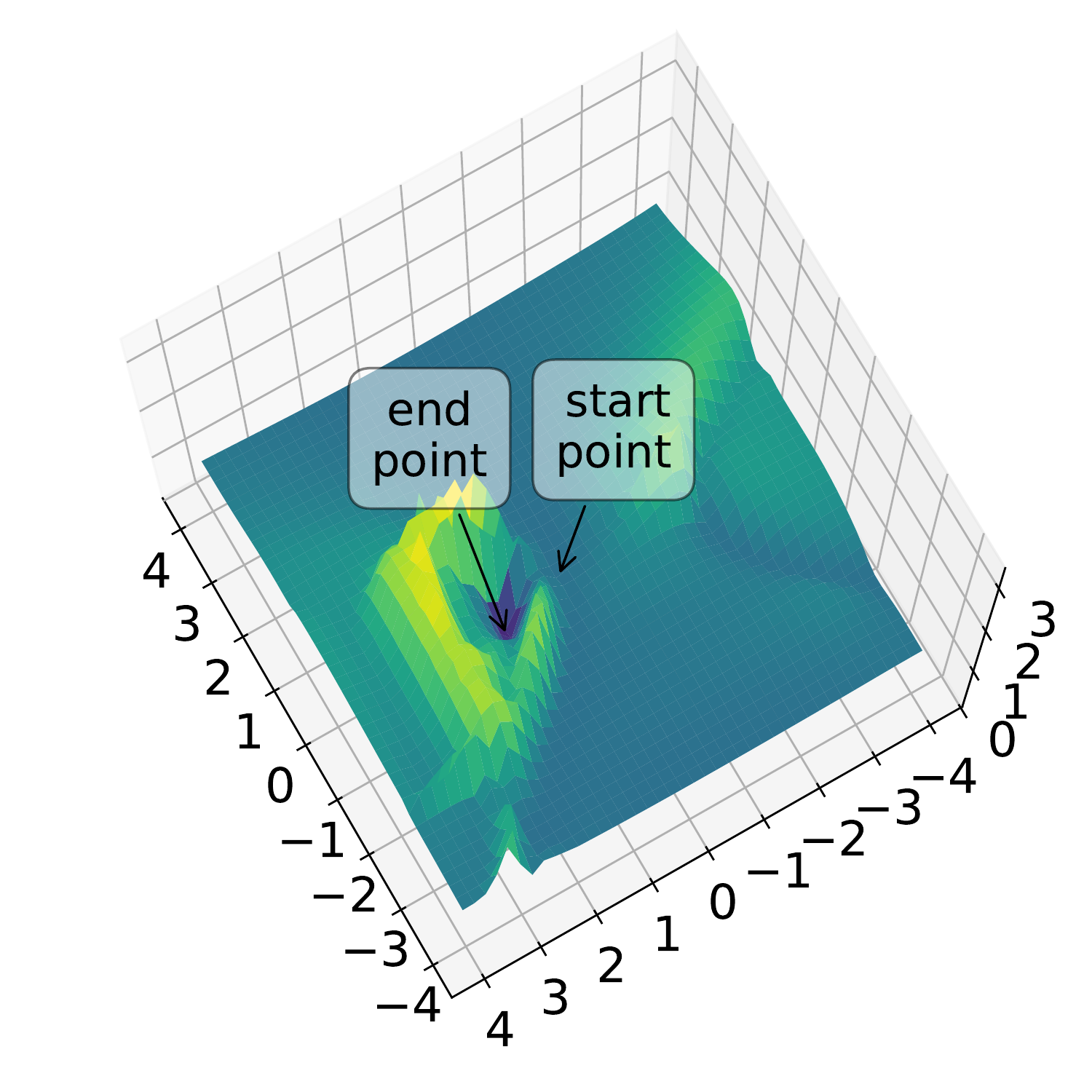}}&
\raisebox{-0.5\height}{\includegraphics[width=0.5\columnwidth]{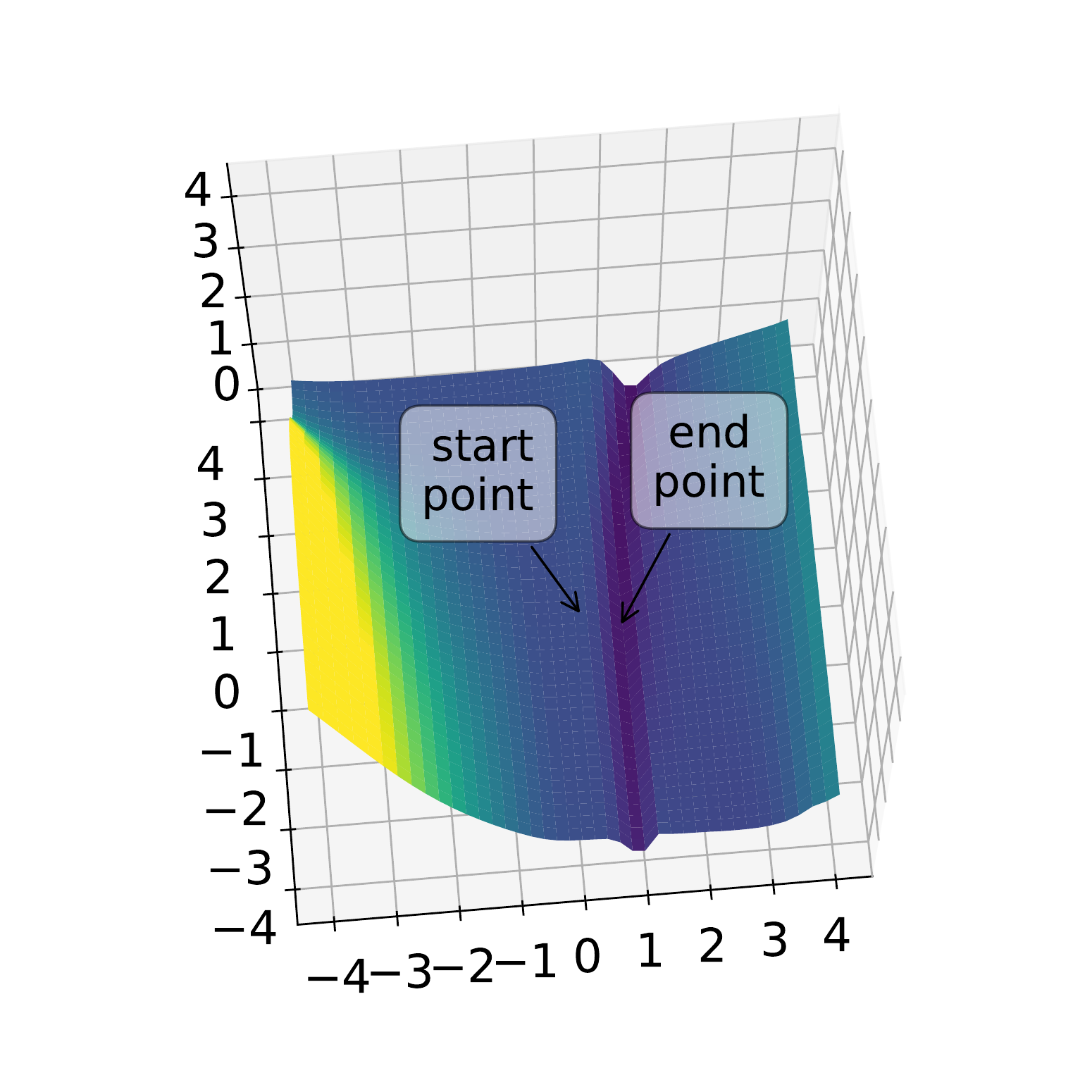}}&
\raisebox{-0.5\height}{\includegraphics[width=0.5\columnwidth]{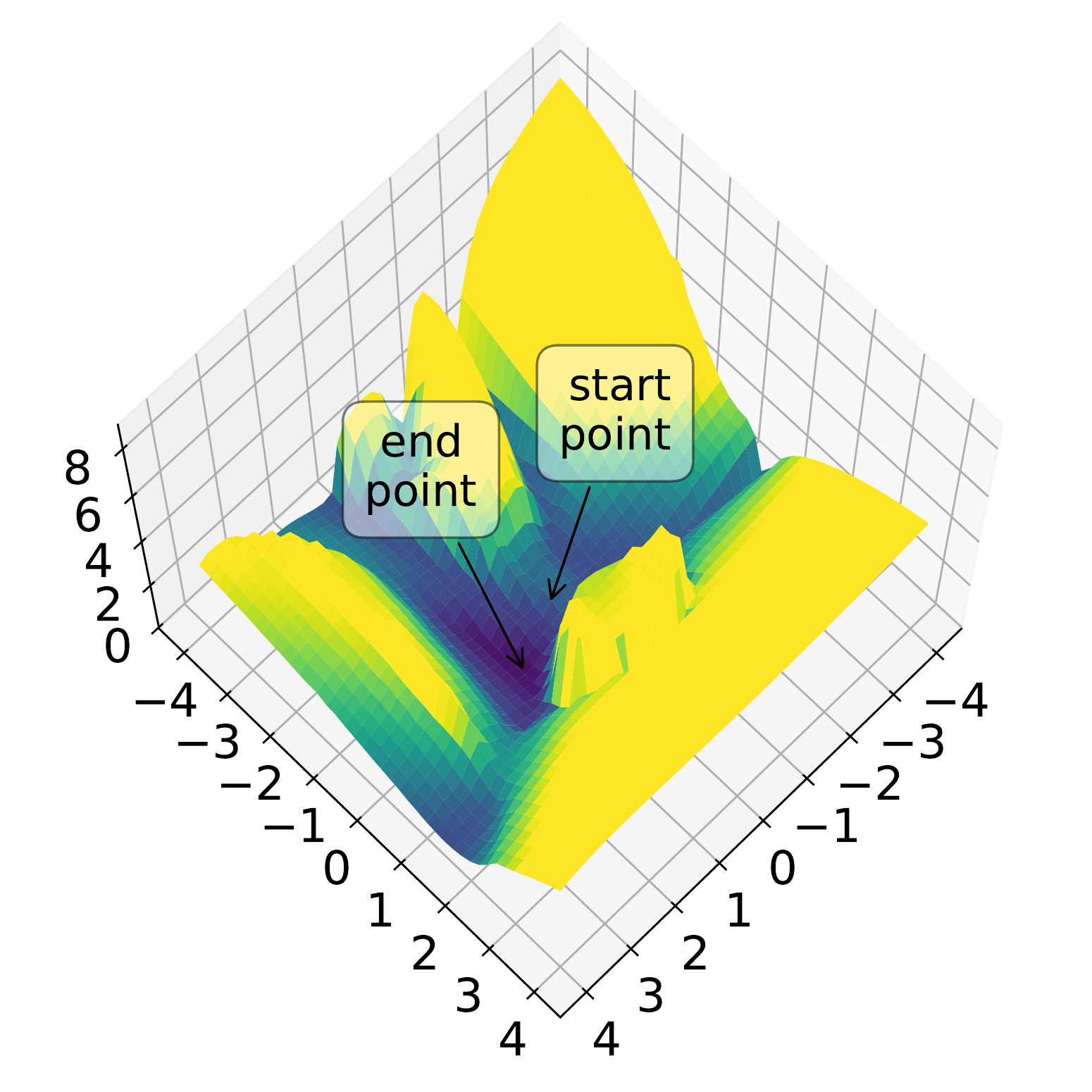}}&
\raisebox{-0.5\height}{\includegraphics[width=0.5\columnwidth]{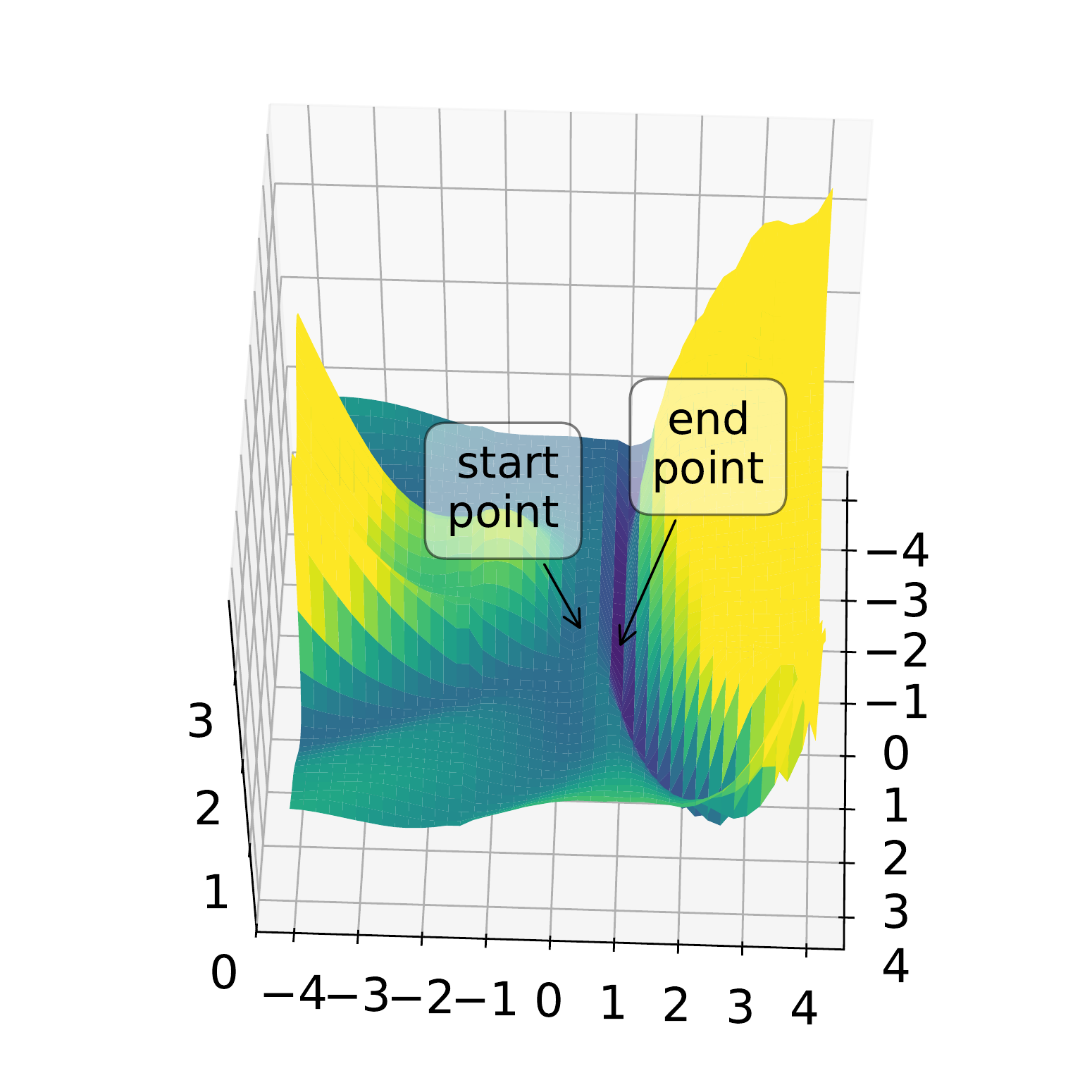}}\\ 
\rotatebox[origin=c]{90}{Fine-tuning BERT}&
\raisebox{-0.5\height}{\includegraphics[width=0.5\columnwidth]{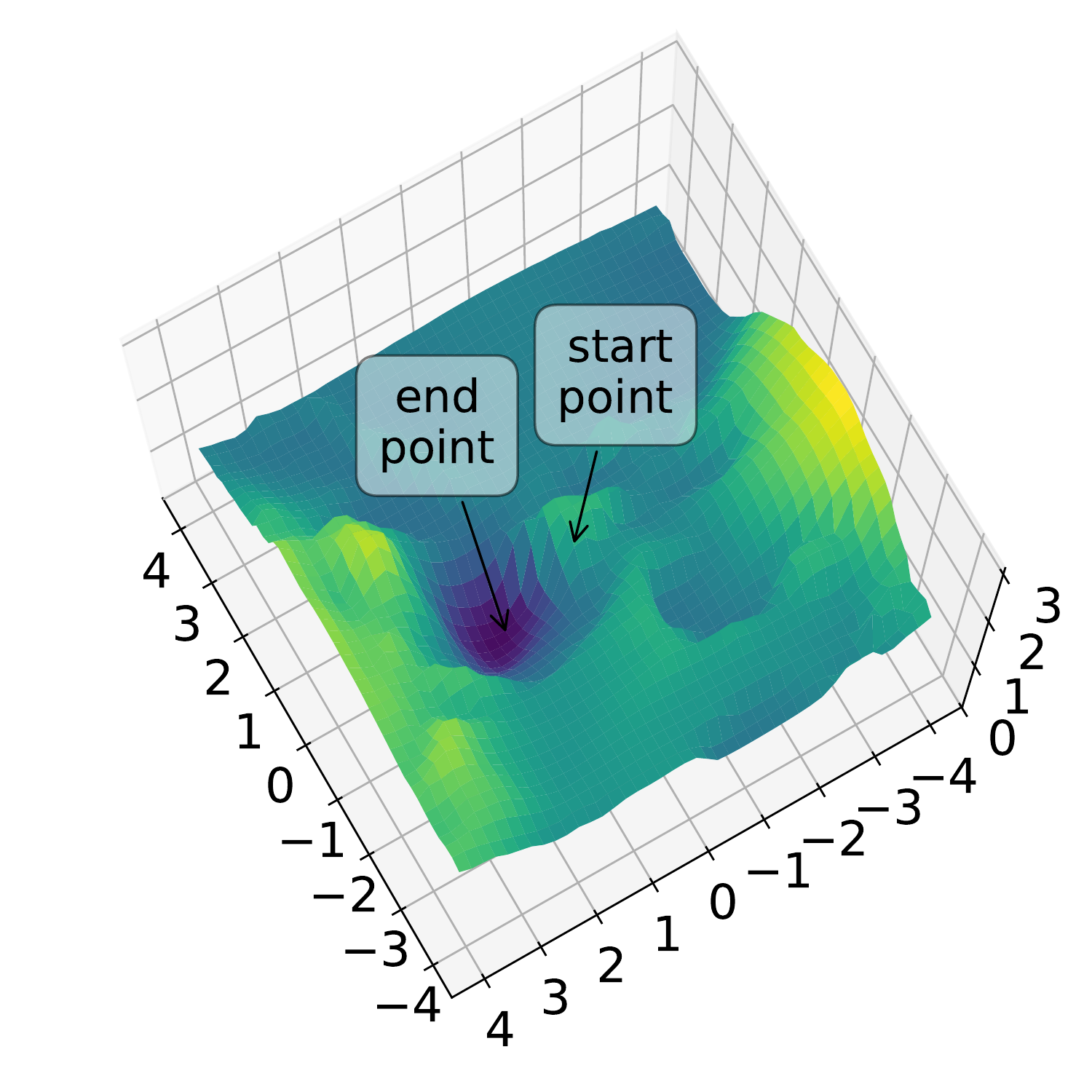}}&
\raisebox{-0.5\height}{\includegraphics[width=0.5\columnwidth]{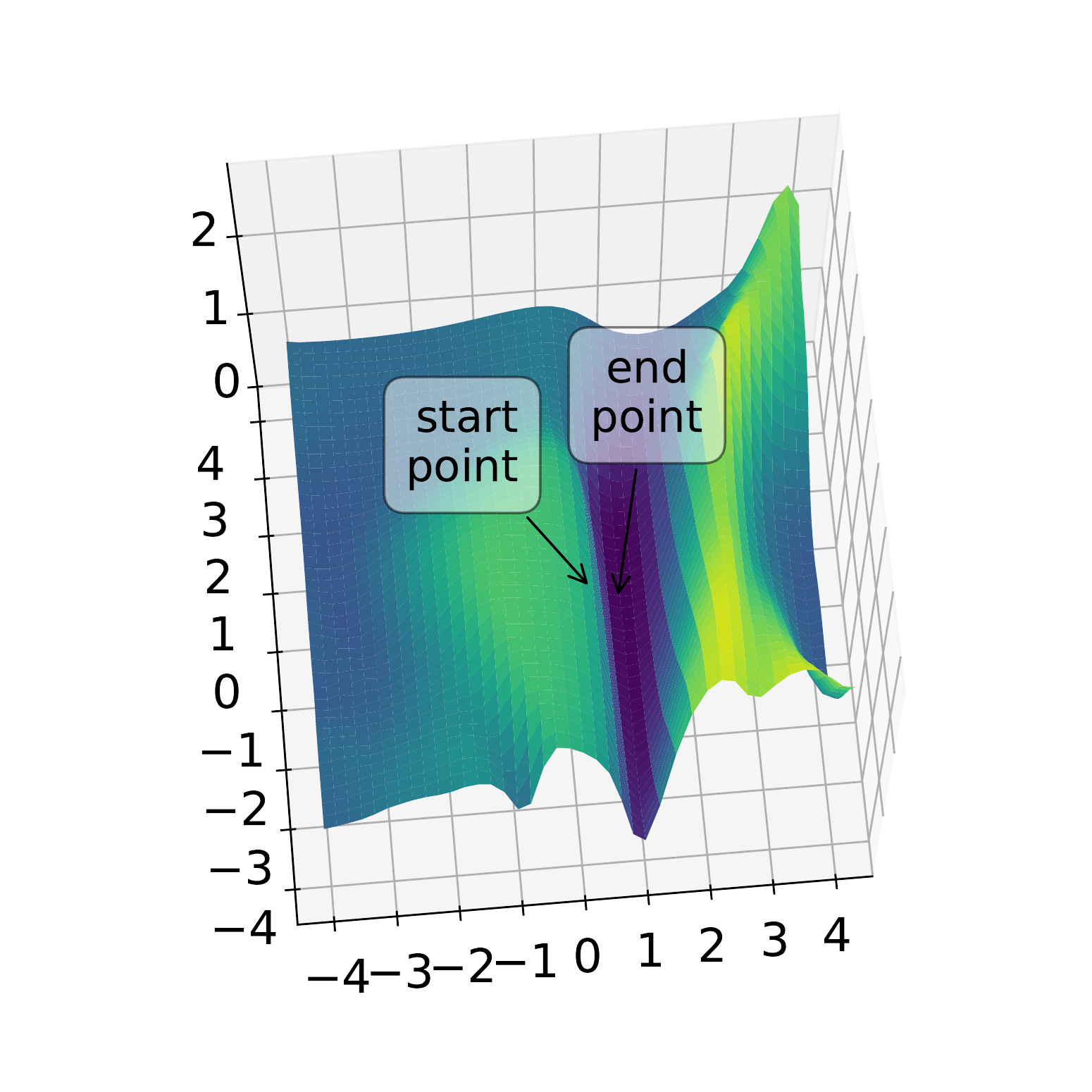}}&
\raisebox{-0.5\height}{\includegraphics[width=0.5\columnwidth]{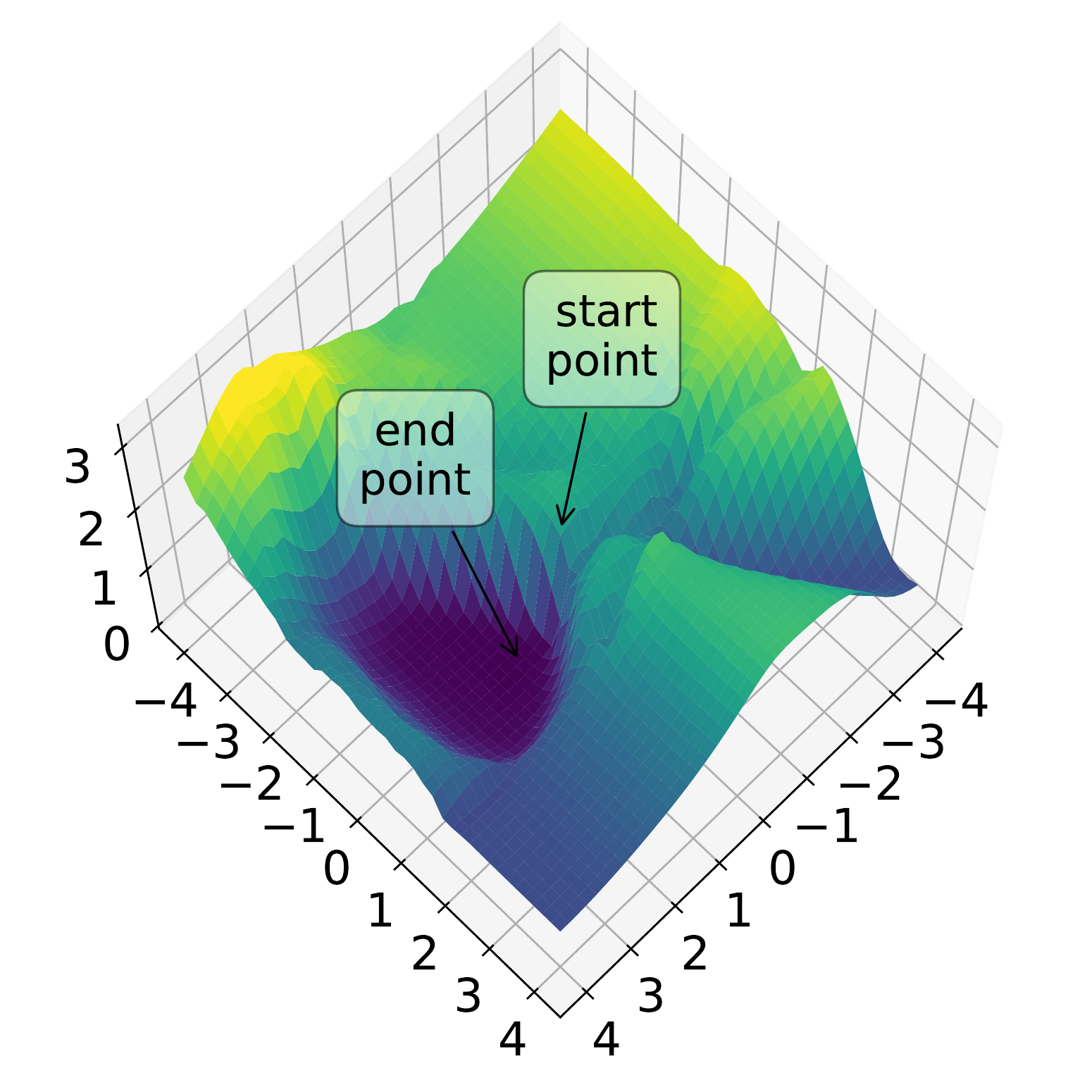}}&
\raisebox{-0.5\height}{\includegraphics[width=0.5\columnwidth]{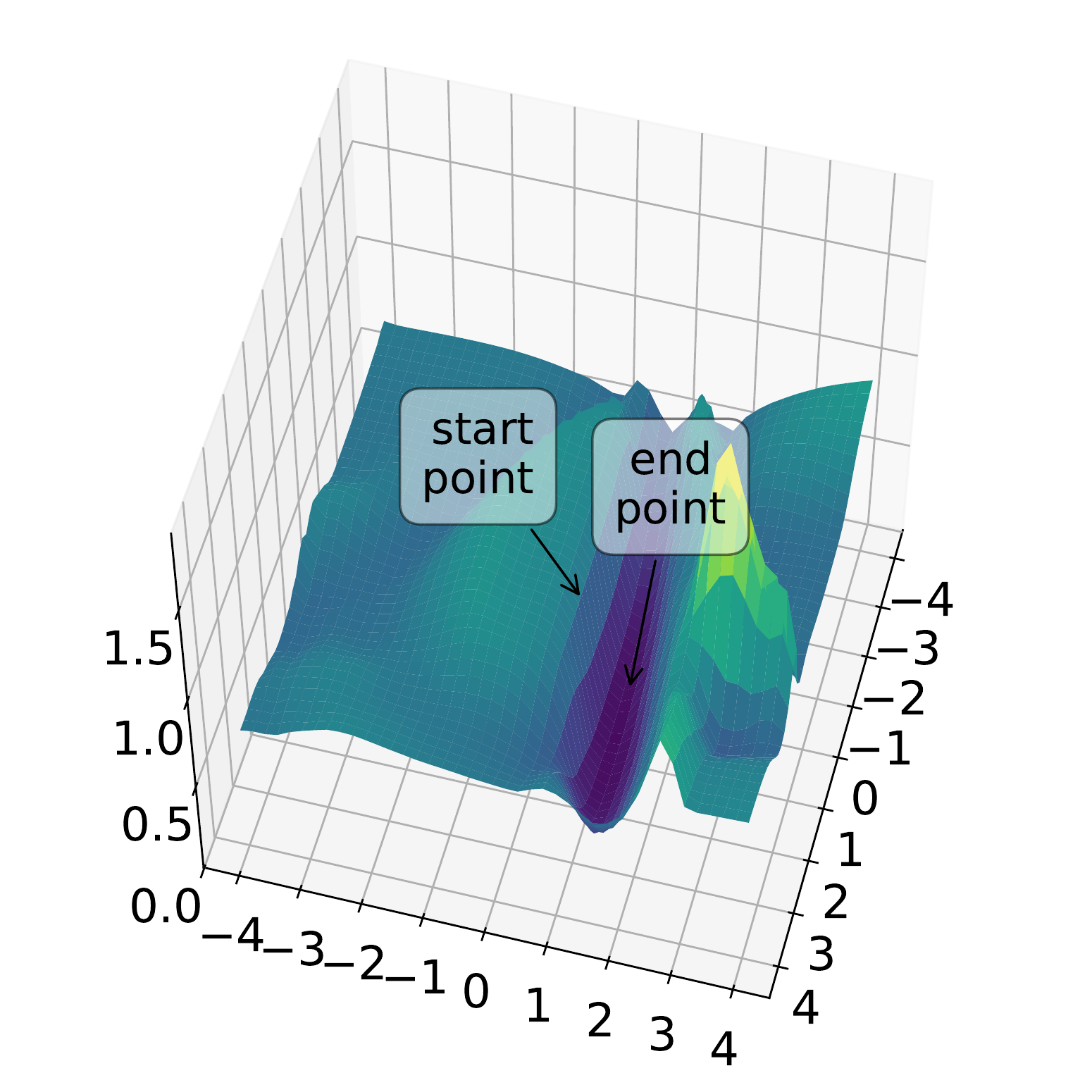}}\\ 
&
\includegraphics[width=0.3\columnwidth]{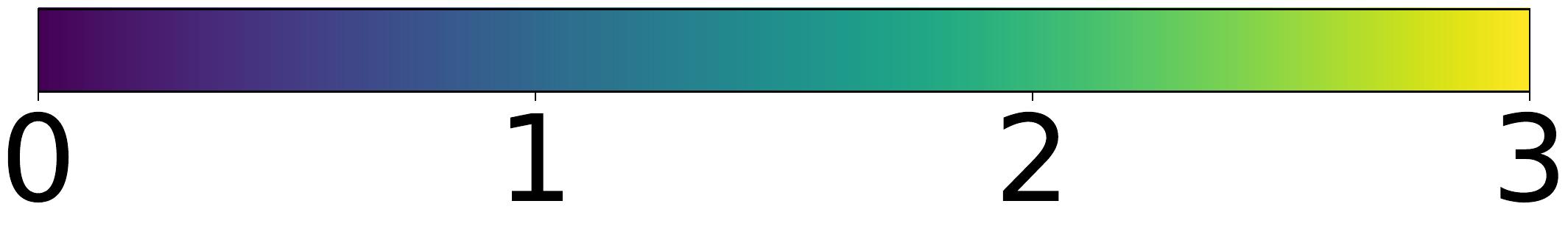}&
\includegraphics[width=0.3\columnwidth]{colorbar0-3.pdf}&
\includegraphics[width=0.3\columnwidth]{colorbar0-3.pdf}&
\includegraphics[width=0.3\columnwidth]{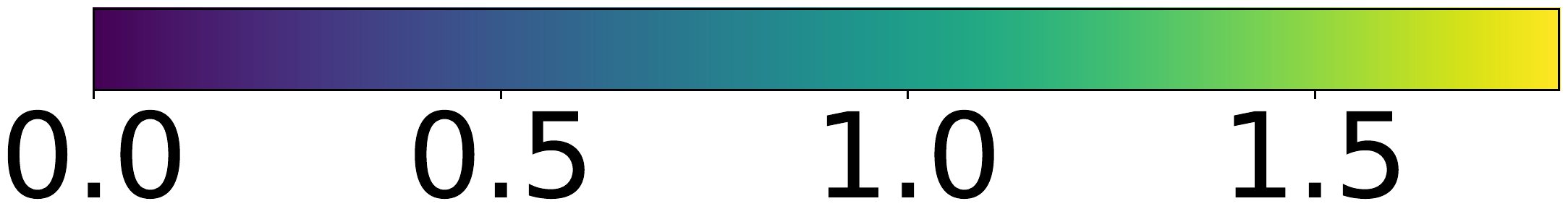}\\ 
\end{tabular}
\caption{Training loss surfaces of training from scratch (top) and fine-tuning BERT (bottom) on four datasets. We annotate the start point (i.e., initialized model) and the end point (i.e., estimated model) in the loss surfaces. Pre-training leads to wider optima, and eases optimization compared with random initialization.}
\label{fig:loss_surface} 
\end{figure*}

As described in Section~\ref{subsc:loss_surface}, we plot 2D training loss surfaces on four datasets in Figure~\ref{fig:loss_surface}.
The figures in the top row are the models trained from scratch, while the others in bottom are based on fine-tuning BERT.
The start point indicates the training loss of the initialized model, while the end point indicates the final training loss.
We observe that the optima obtained by fine-tuning BERT are much wider than training from scratch.
A wide optimum of fine-tuning BERT implicates that the small perturbations of the model parameters cannot hurt the final performance seriously, while a thin optimum is more sensitive to these subtle changes.
Moreover, in Section~\ref{sub:wide:generalize}, we further discuss about the width of the optima can contribute to the generalization capability.

As shown in Figure~\ref{fig:loss_surface}, the fine-tuning path from the start point to the end point on the loss landscape is more smooth than training from scratch. In other words, the training loss of fine-tuning BERT tends to monotonously decrease along the optimization direction, which eases optimization and accelerates training convergence.
In contrast, the path from random initial point to the end point is more rough, which requires a more carefully tweaked optimizer to obtain reasonable performance.

\subsection{Pre-training Eases Optimization on Downstream Tasks}

\begin{figure}[t]
\centering
\includegraphics[width=0.48\textwidth]{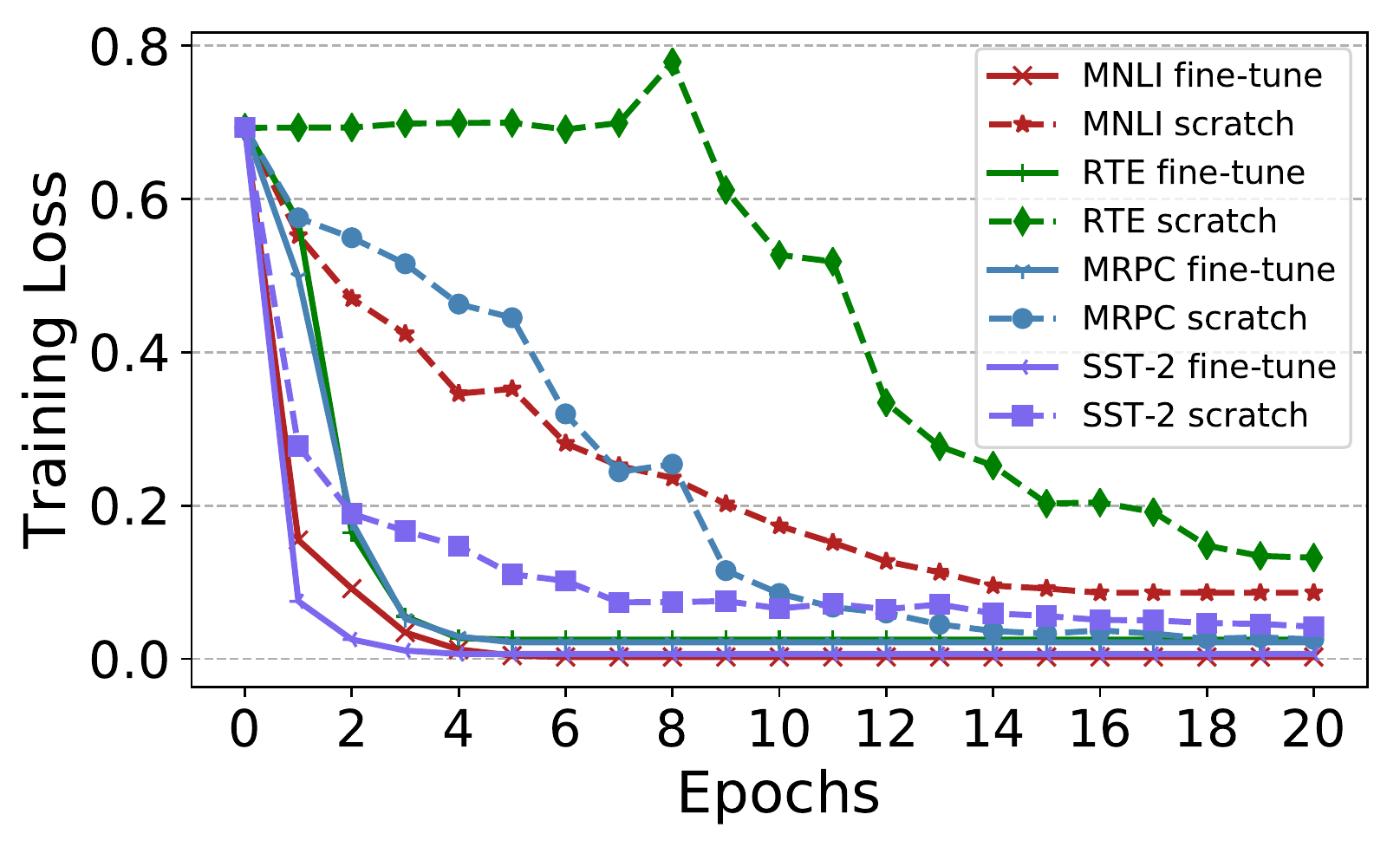}
\caption{Training loss of fine-tuning BERT and training from scratch on four datasets.}
\label{fig:loss_curve}
\end{figure}

We fine-tune BERT and train the same network from scratch on four datasets. The learning curves are shown in Figure~\ref{fig:loss_curve}.
We find that training from scratch requires more iterations to converge on the datasets, while pre-training-then-fine-tuning converges faster in terms of training loss.
We also notice that the final loss of training from scratch tends to be higher than fine-tuning BERT, even if it undergoes more epochs. On the RTE dataset, training the model from scratch has a hard time decreasing the loss in the first few epochs.

\begin{figure}[t]
\centering
\small
\begin{tabular}{l@{\hspace*{\lengtha}}c@{\hspace*{\lengtha}}c@{\hspace*{\lengtha}}c}
&MRPC&MNLI\\
\rotatebox[origin=c]{90}{Training from scratch}&
\raisebox{-0.5\height}{\includegraphics[width=0.46\columnwidth]{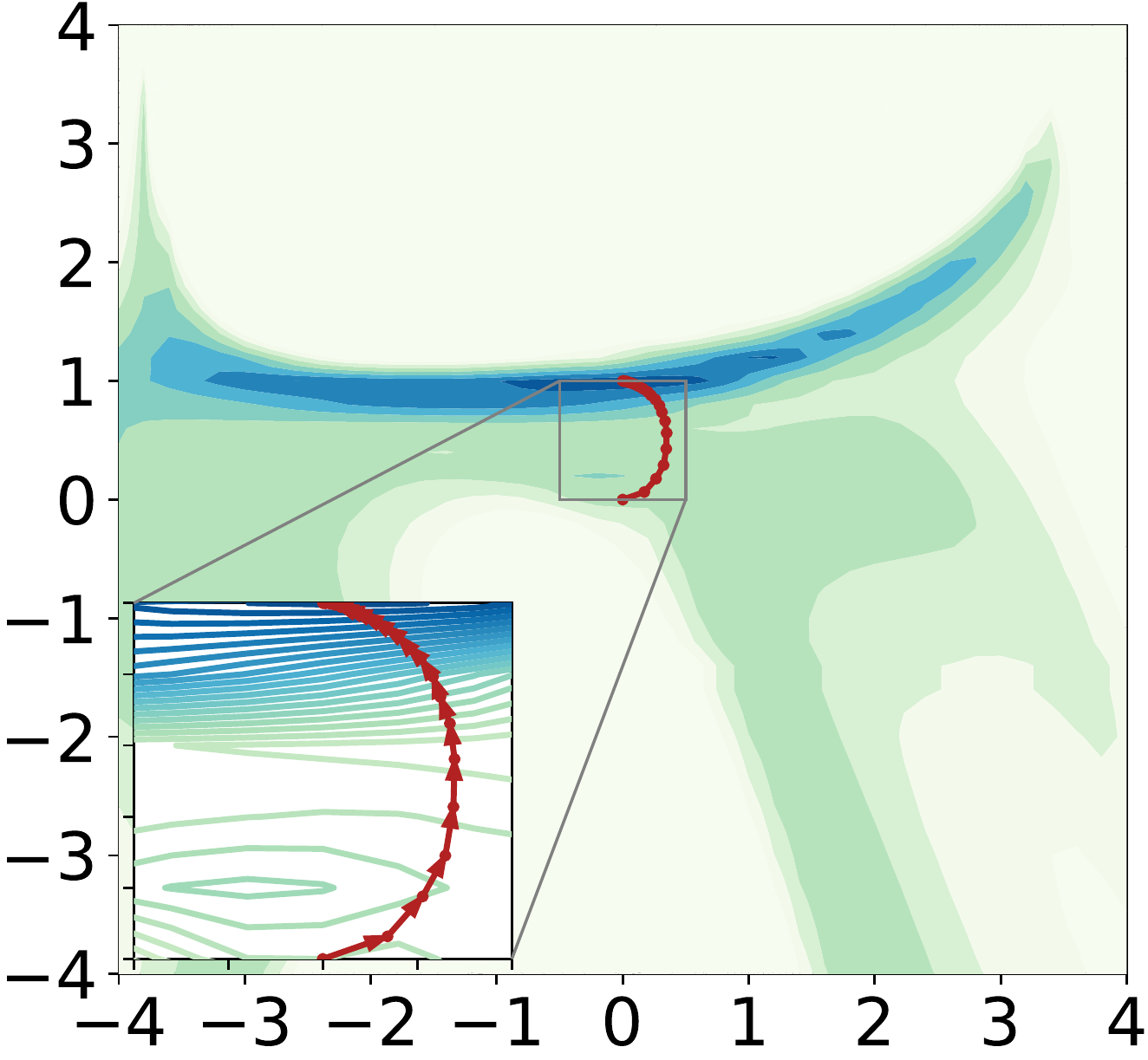}}&
\raisebox{-0.5\height}{\includegraphics[width=0.46\columnwidth]{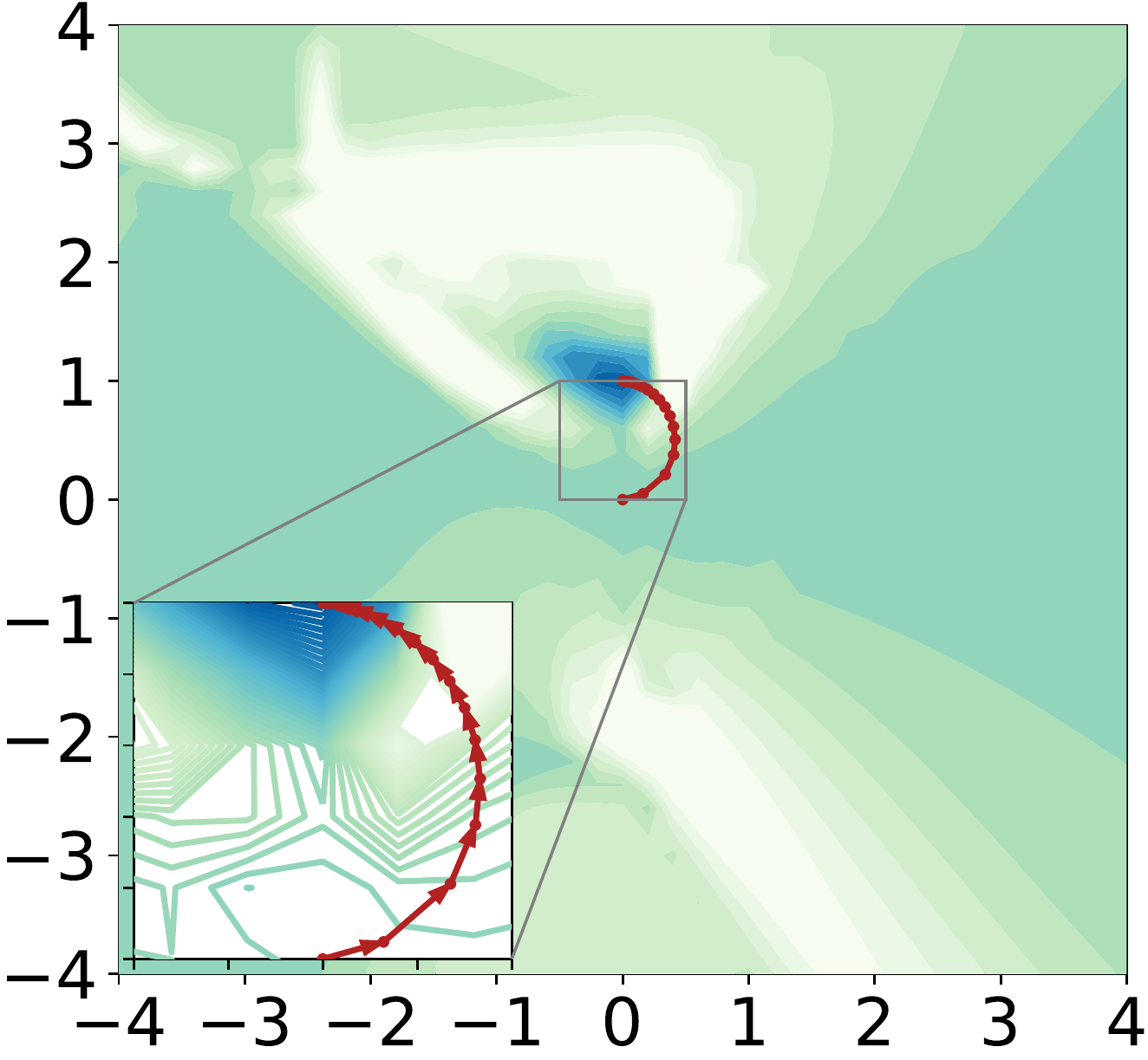}}\\
\addlinespace[0.3cm]
\rotatebox[origin=c]{90}{Fine-tuning BERT}&
\raisebox{-0.5\height}{\includegraphics[width=0.46\columnwidth]{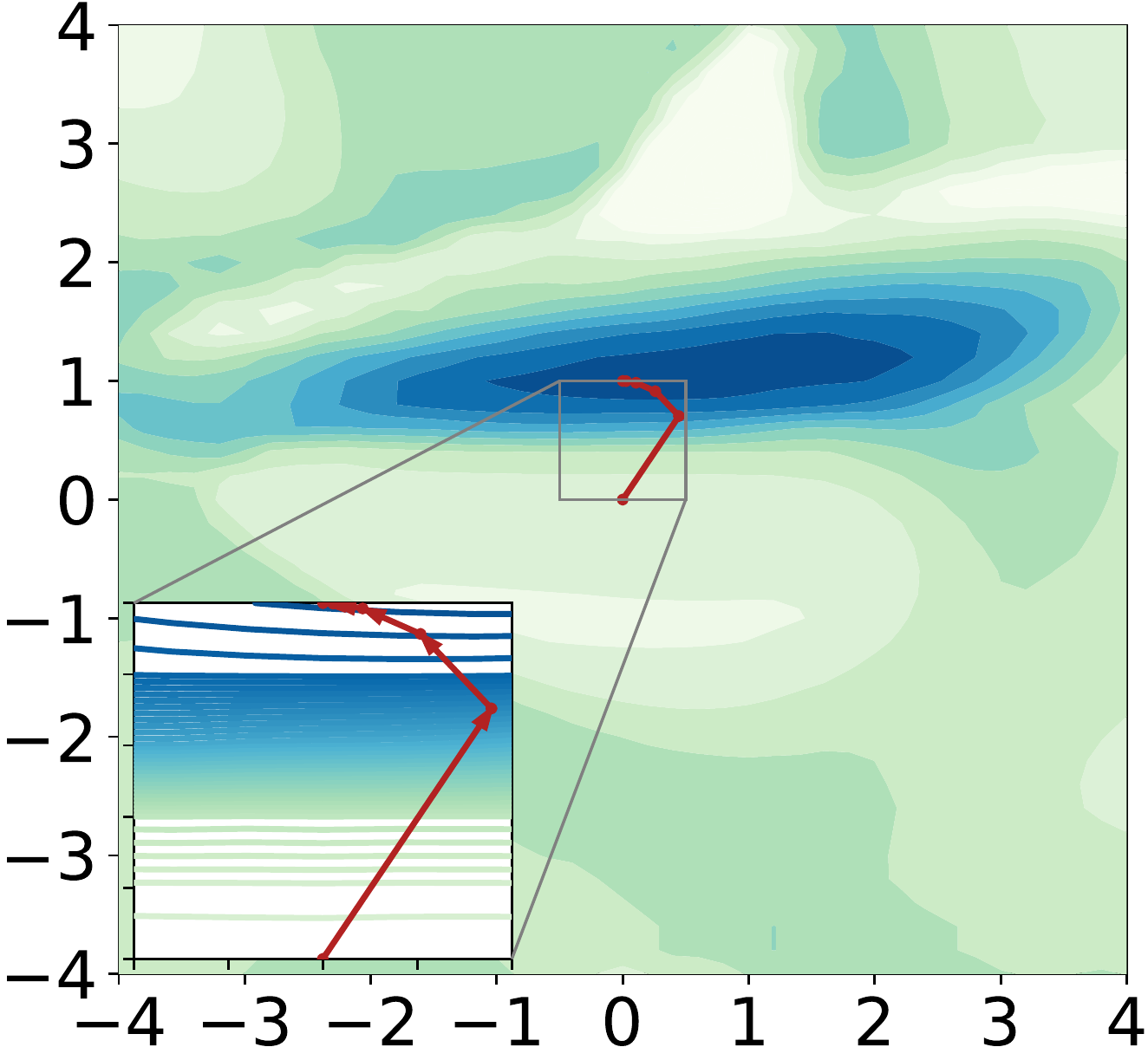}}&
\raisebox{-0.5\height}{\includegraphics[width=0.46\columnwidth]{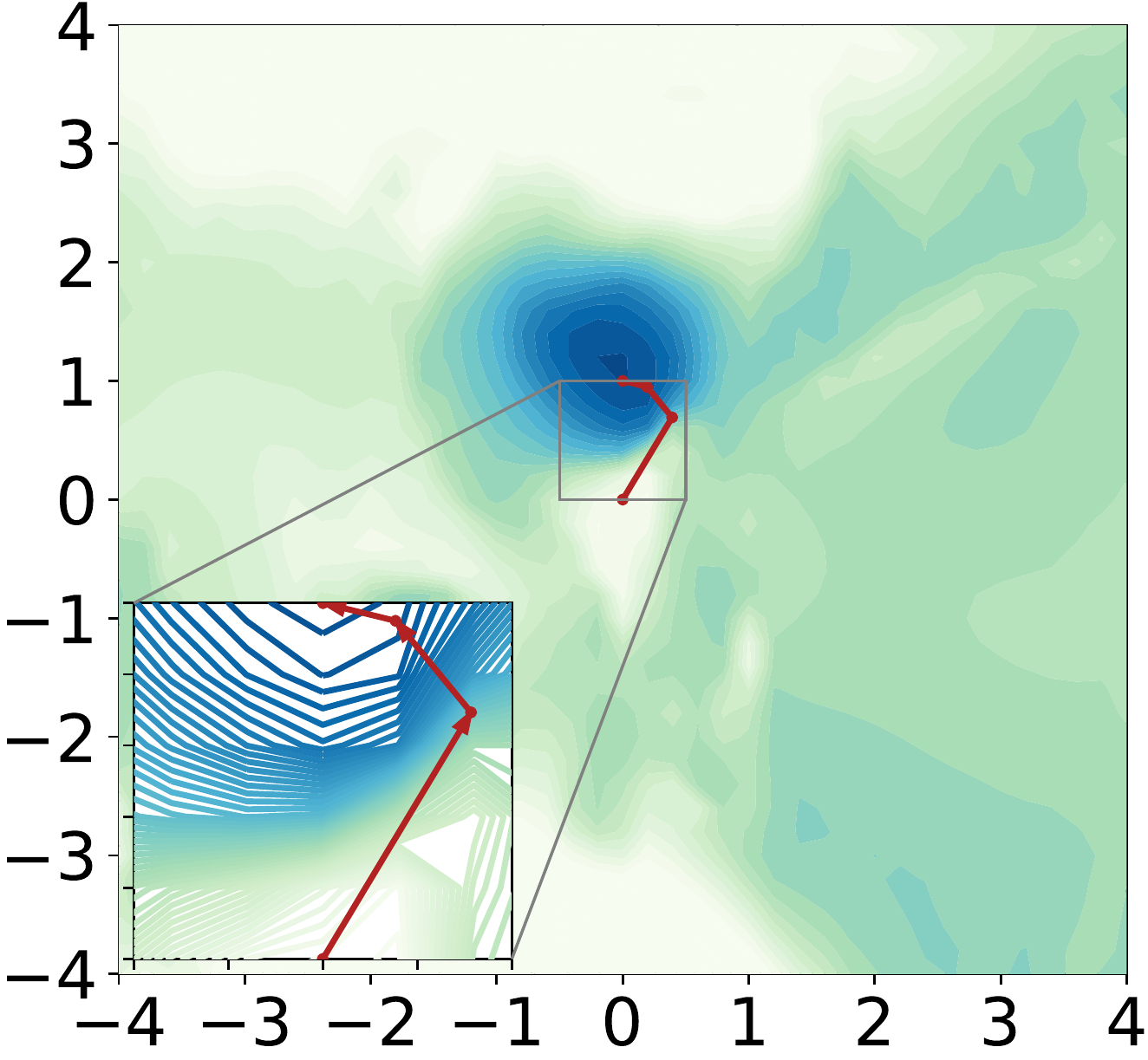}}\\
\addlinespace[2mm]
&
\includegraphics[width=0.3\columnwidth]{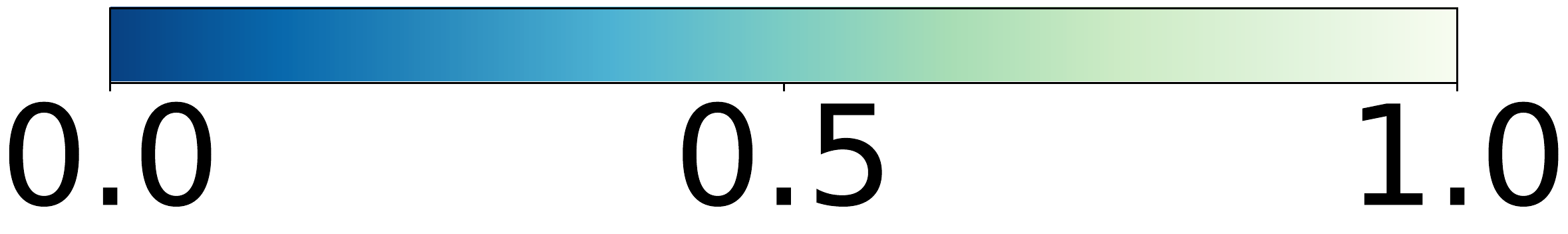}&
\includegraphics[width=0.3\columnwidth]{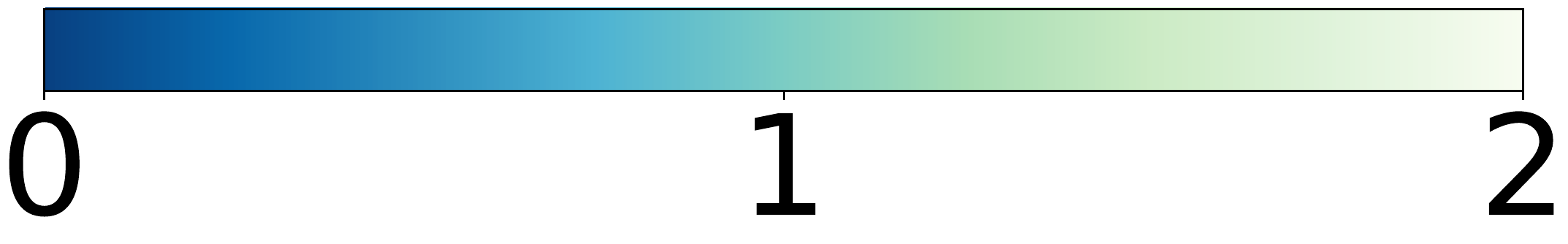}\\
\end{tabular}
\caption{The optimization trajectories on the training loss surfaces of training from scratch (top) and fine-tuning BERT (bottom) on the MRPC and MNLI datasets. Darker color represents smaller training loss.}
\label{fig:loss_traj} 
\end{figure}

In order to visualize the dynamic convergence process, we plot the optimization trajectories using the method described in Section~\ref{subsc:opt_traj}.
As shown in Figure~\ref{fig:loss_traj}, for training from scratch, the optimization directions of the first few epochs are divergent from the final optimization direction. Moreover, the loss landscape from the initial point to the end point is more rough than fine-tuning BERT, we can see that the trajectory of training from scratch on the MRPC dataset crosses an obstacle to reach the end point.

Compared with training from scratch, fine-tuning BERT finds the optimization direction in a more straightforward way. The optimization process also converges faster.
Besides, the fine-tuning path is unimpeded along the optimization direction.
In addition, because of the wider optima near the initial point, fine-tuning BERT tends to reach the expected optimal region even if it optimizes along the direction of the first epoch.


\subsection{Pre-training-then-fine-tuning is Robust to Overfitting}

The BERT-large model has 345M parameters, which is over-parameterized for the target datasets.
However, experimental results show fine-tuning BERT is robust to over-fitting, i.e., the generalization error (namely, the classification error rate on the development set) does not dramatically increase for more training epochs, despite the huge number of model parameters.

\begin{figure}[t]
\centering
\includegraphics[width=0.46\textwidth]{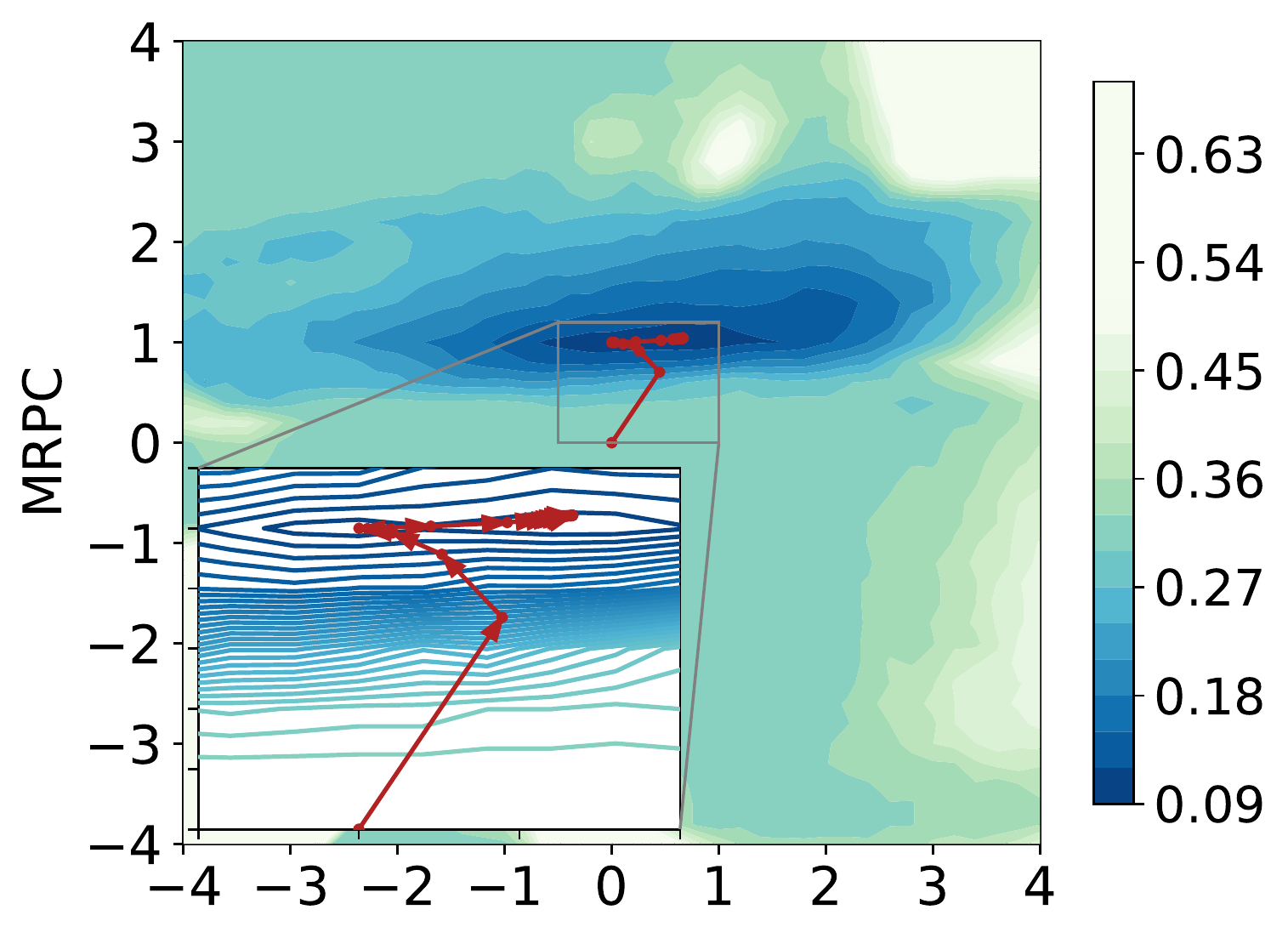}
\caption{The optimization trajectory of fine-tuning $5+20$ epochs on MRPC. The two-dimensional generalization error surface is presented. We find that pre-training-then-fine-tuning is robust to overfitting.}
\label{fig:O-mrpc-traj-rt20}
\end{figure}

We use the MRPC dataset as a case study, because its data size is relatively small, which is prone to overfitting if we train the model from scratch.
As shown in Figure~\ref{fig:O-mrpc-traj-rt20}, we plot the optimization trajectory of fine-tuning on the generalization error surface.
We first fine-tune the BERT model for five epochs as suggested in~\cite{bert}. Then we continue fine-tuning for another twenty epochs, which still obtains comparable performance with the first five epochs.
Figure~\ref{fig:O-mrpc-traj-rt20} shows that even though we fine-tune the BERT model for twenty more epochs, the final estimation is not far away from its original optimum.
Moreover, the optimum area is wide enough to avoid the model from jumping out the region with good generalization capability, which explains why the pre-training-then-fine-tuning paradigm is robust to overfitting.

\section{Pre-training Helps to Generalize Better}

Although training from scratch can achieve comparable training losses as fine-tuning BERT, the model with random initialization usually has poor performance on the unseen data.
In this section, we use visualization techniques to understand why the model obtained by pre-training-then-fine-tuning tends to have better generalization capability.

\subsection{Wide and Flat Optima Lead to Better Generalization}
\label{sub:wide:generalize}

Previous work~\cite{flatminima,sharpness,visualloss} shows that the flatness of a local optimum correlates with the generalization capability, i.e., more flat optima lead to better generalization.
The finding inspires us to inspect the loss landscapes of BERT fine-tuning, so that we can understand the generalization capability from the perspective of the flatness of optima.


\begin{figure}[t]
\centering
\small
\begin{tabular}{c@{\hspace*{\lengtha}}c@{\hspace*{\lengtha}}c}
\raisebox{-0.5\height}{\includegraphics[width=0.48\columnwidth]{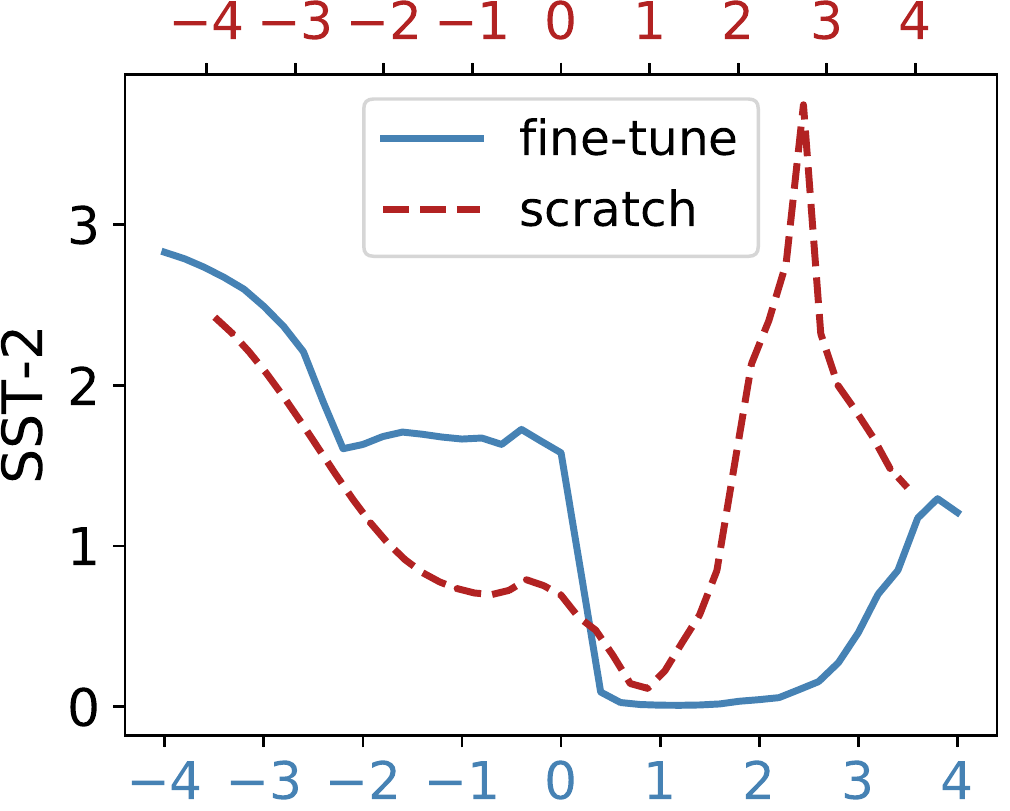}}&
\raisebox{-0.5\height}{\includegraphics[width=0.48\columnwidth]{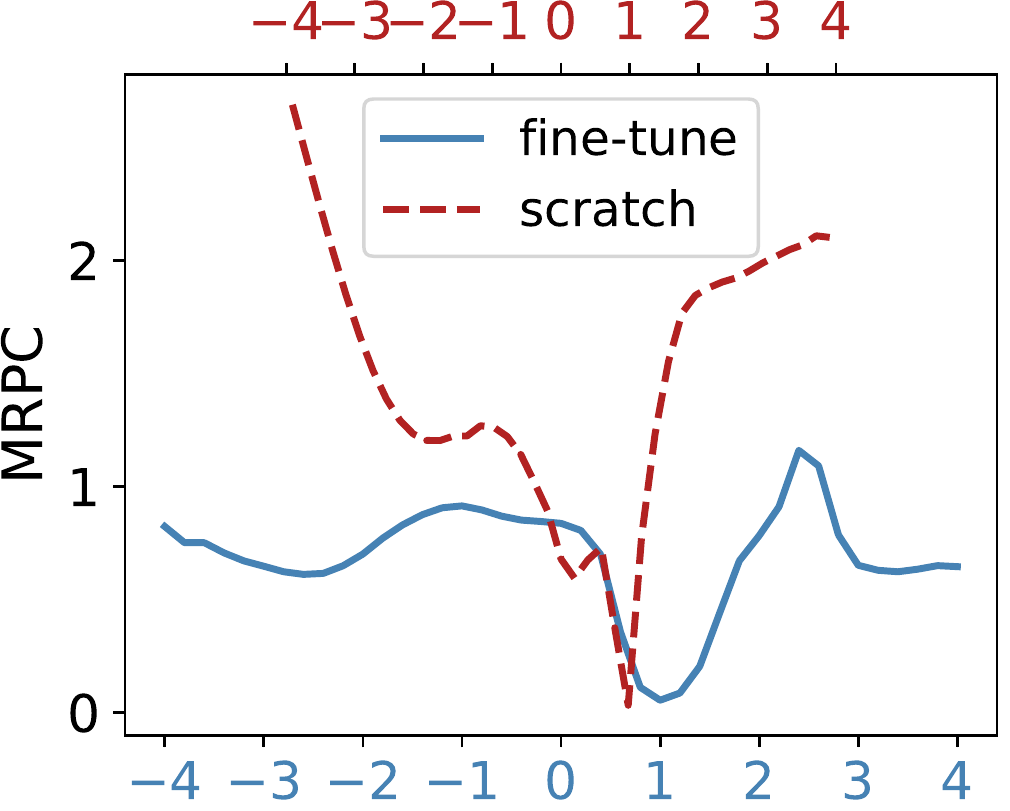}}\\
\addlinespace[0.3cm]
\raisebox{-0.5\height}{\includegraphics[width=0.48\columnwidth]{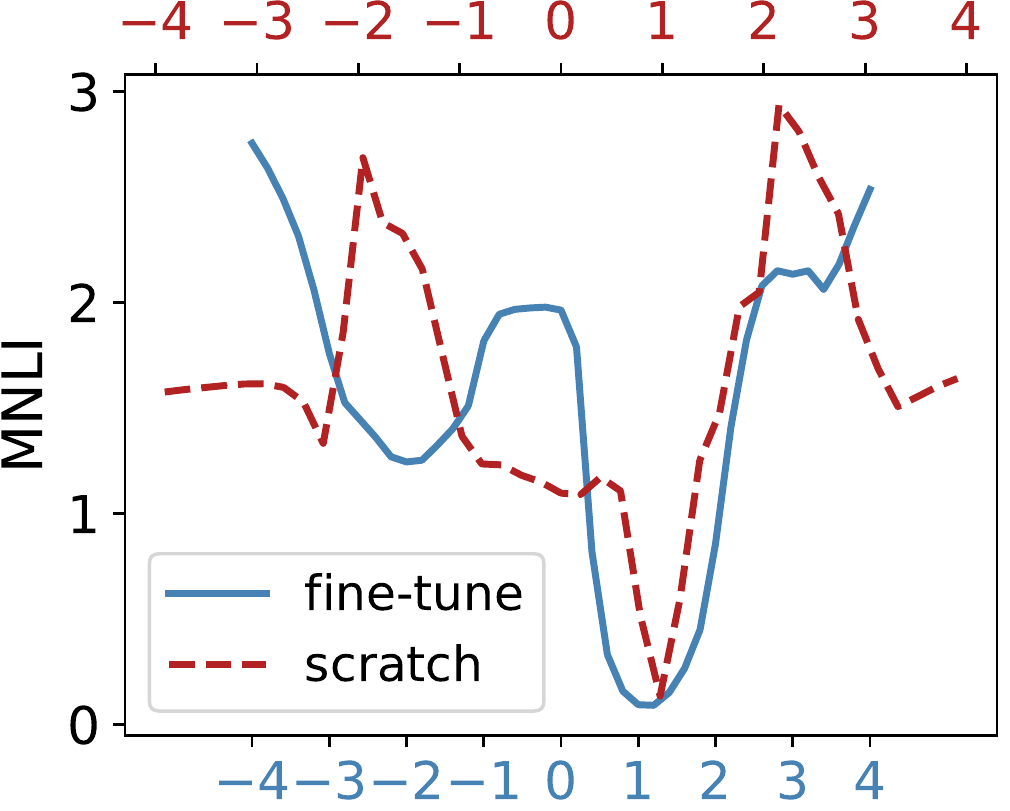}}&
\raisebox{-0.5\height}{\includegraphics[width=0.48\columnwidth]{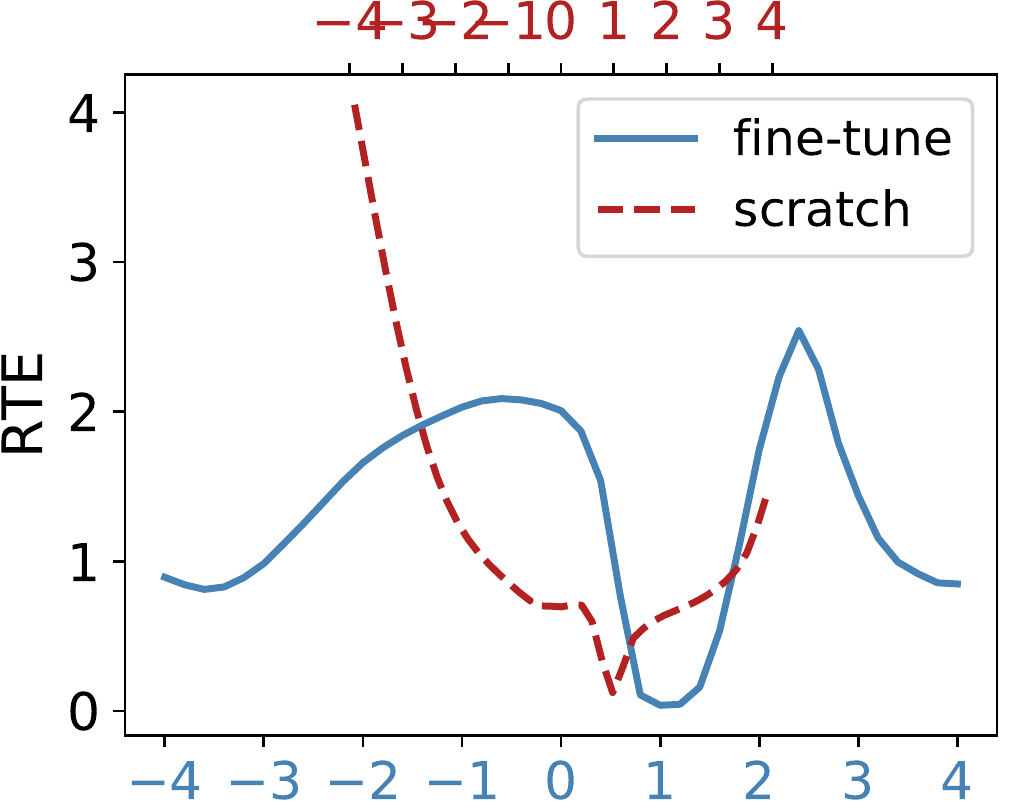}}\\
\end{tabular}
\caption{One-dimensional training loss curves. Dash lines represent training from scratch, and solid lines represent fine-tuning BERT. The scales of axes are normalized for comparison. The optima of training from scratch are sharper than fine-tuning BERT.}
\label{fig:sharpness} 
\end{figure}

Section~\ref{subsc:wider_optima} presents that the optima obtained by fine-tuning BERT are wider than training from scratch.
As shown in Figure~\ref{fig:sharpness}, we further plot one-dimensional training loss curves of both fine-tuning BERT and training from scratch, which represents the transverse section of two-dimensional loss surface along the optimization direction.
We normalize the scale of axes for flatness comparison as suggested in~\cite{visualloss}.
Figure~\ref{fig:sharpness} shows that the optima of fine-tuning BERT are more flat, while training from scratch obtains more sharp optima.
The results indicate that pre-training-then-fine-tuning tends to generalize better on unseen data.

\subsection{Consistency Between Training Loss Surface and Generalization Error Surface}

\begin{figure*}[t]
\centering
\small
\begin{tabular}{l@{\hspace*{\lengtha}}c@{\hspace*{\lengtha}}c@{\hspace*{\lengtha}}c@{\hspace*{\lengtha}}c@{\hspace*{\lengtha}}c}
&MNLI&RTE&SST-2&MRPC\\
\rotatebox[origin=c]{90}{Training from scratch}&
\raisebox{-0.5\height}{\includegraphics[width=0.492\columnwidth]{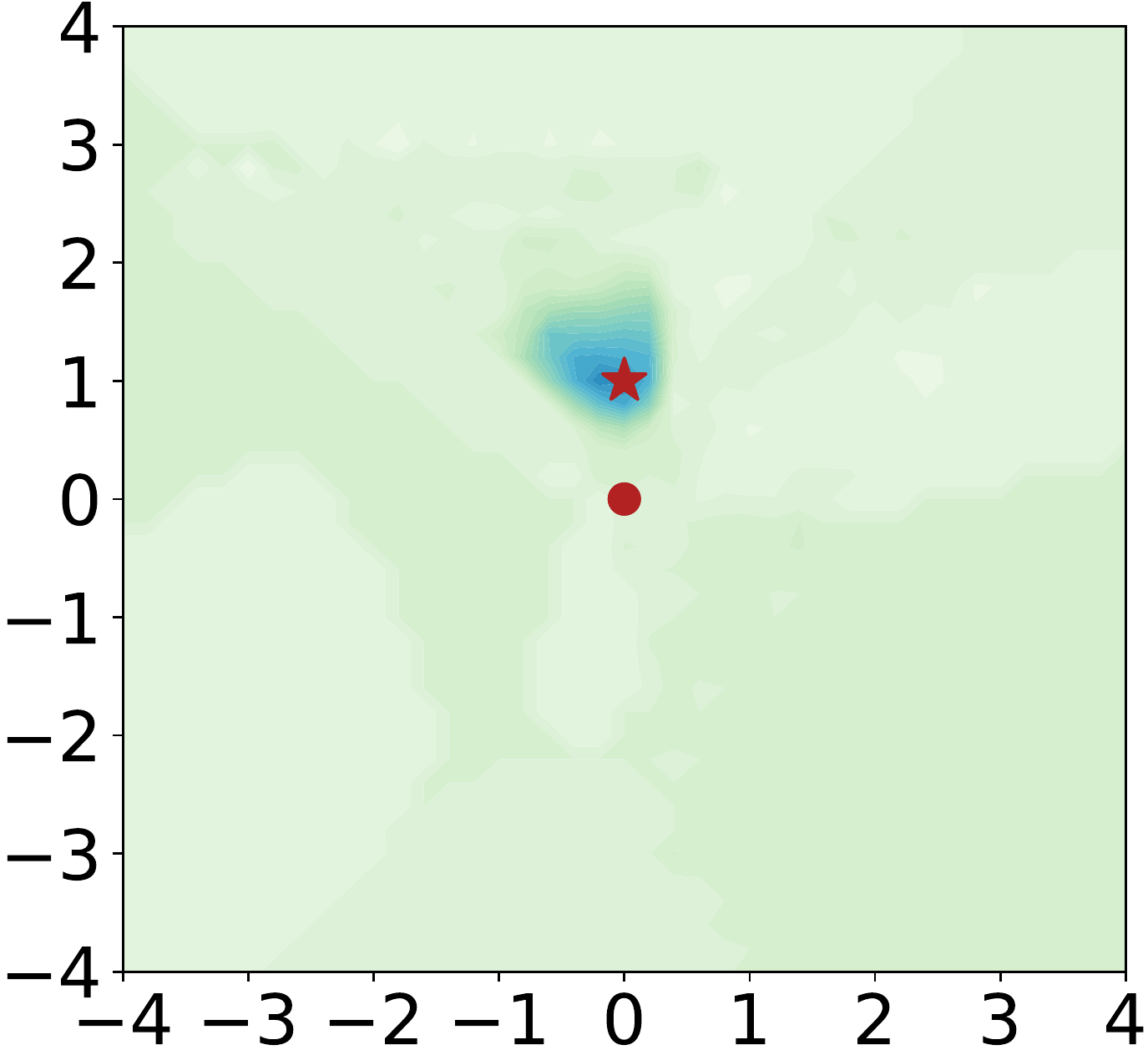}}&
\raisebox{-0.5\height}{\includegraphics[width=0.492\columnwidth]{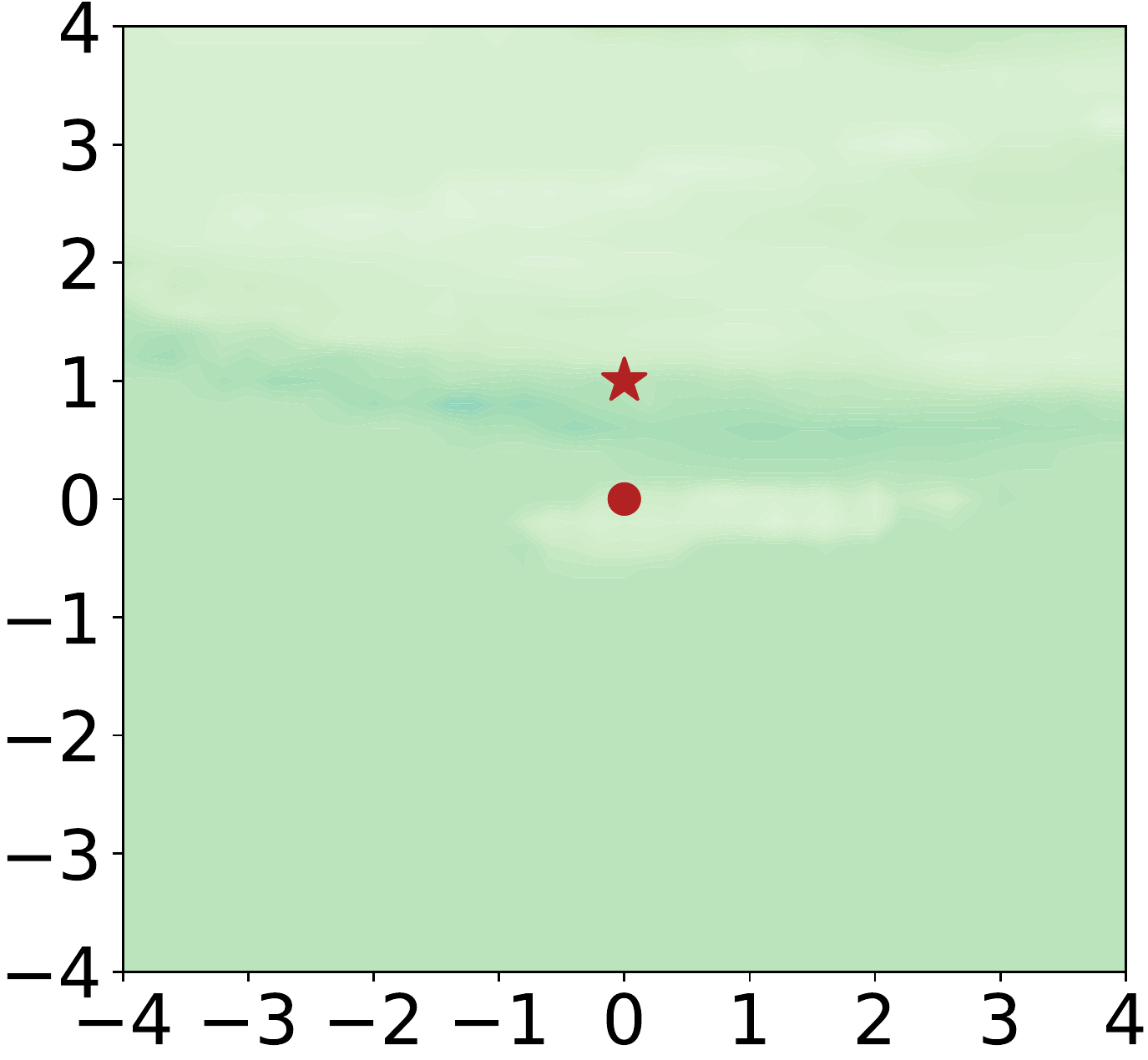}}&
\raisebox{-0.5\height}{\includegraphics[width=0.492\columnwidth]{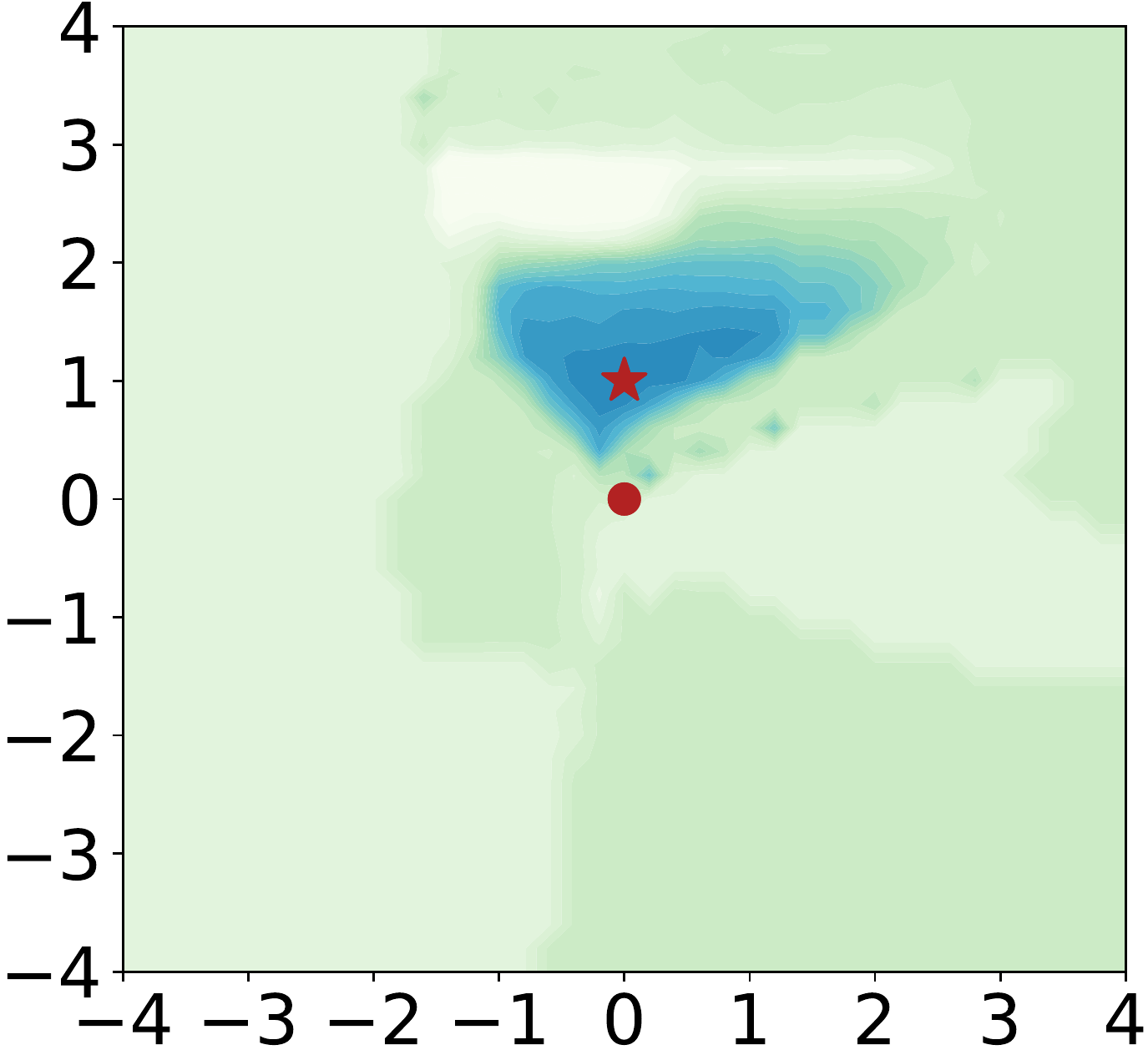}}&
\raisebox{-0.5\height}{\includegraphics[width=0.492\columnwidth]{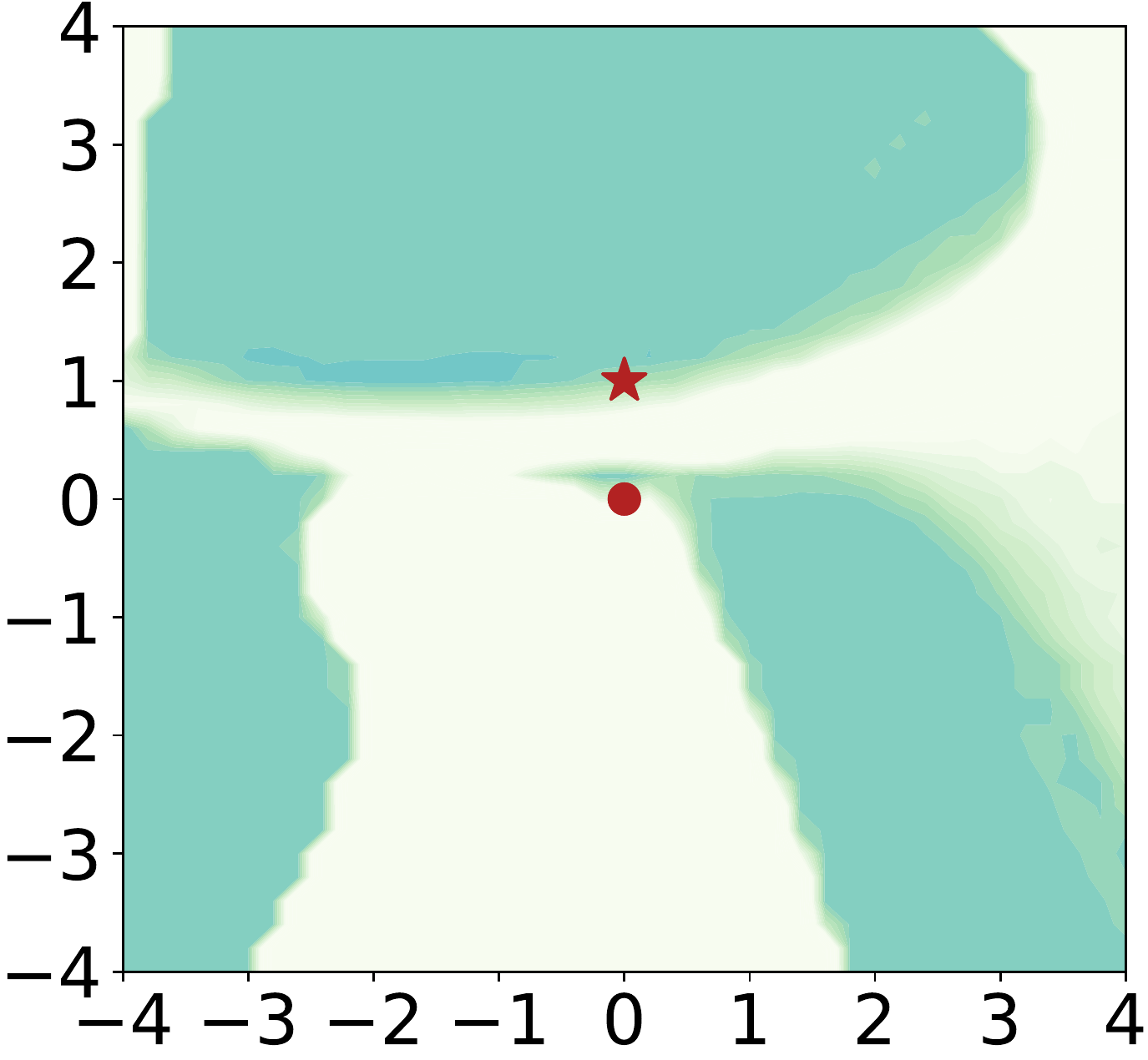}}\\
\addlinespace[0.3cm]
\rotatebox[origin=c]{90}{Fine-tuning BERT}&
\raisebox{-0.5\height}{\includegraphics[width=0.492\columnwidth]{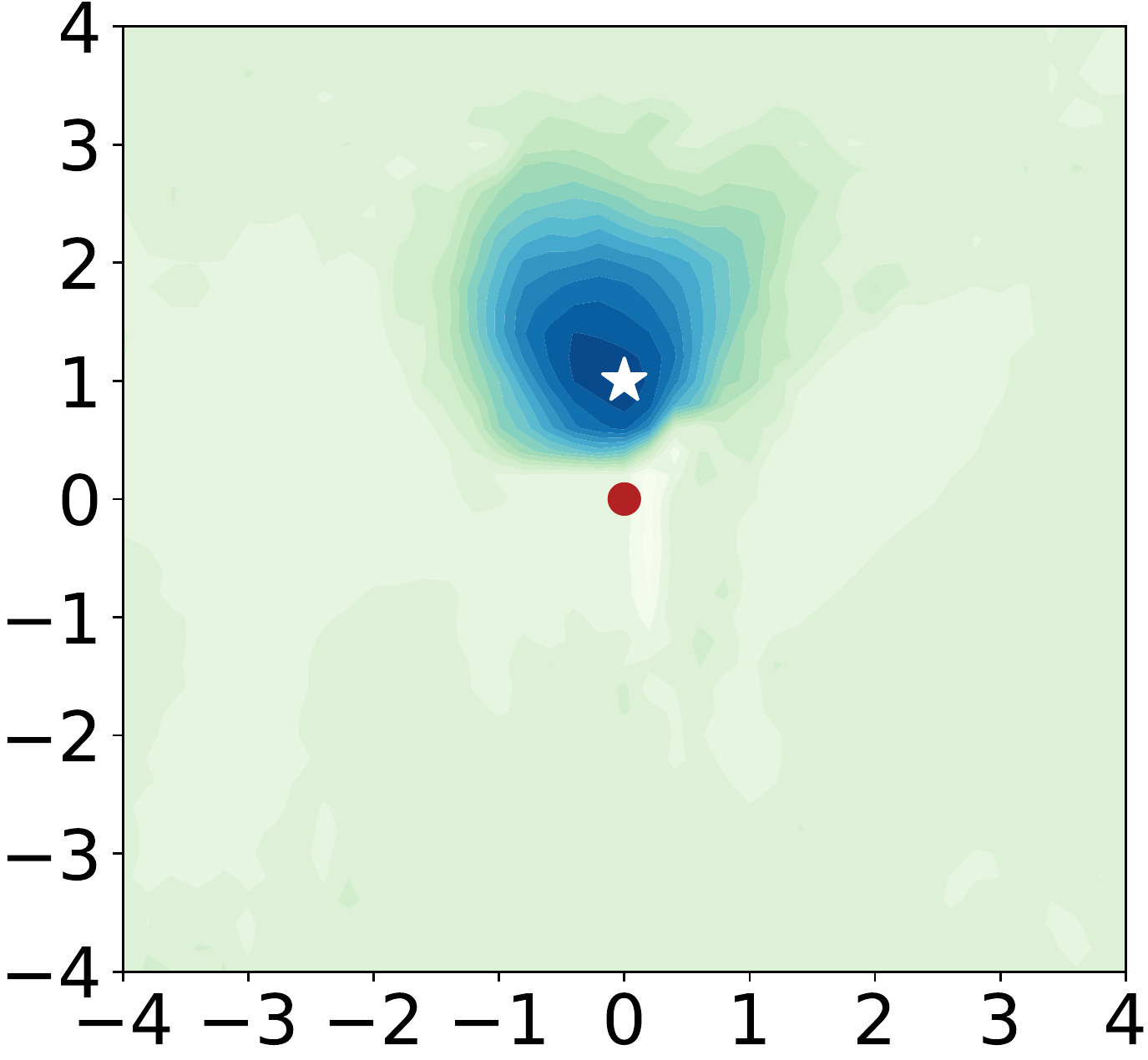}}&
\raisebox{-0.5\height}{\includegraphics[width=0.492\columnwidth]{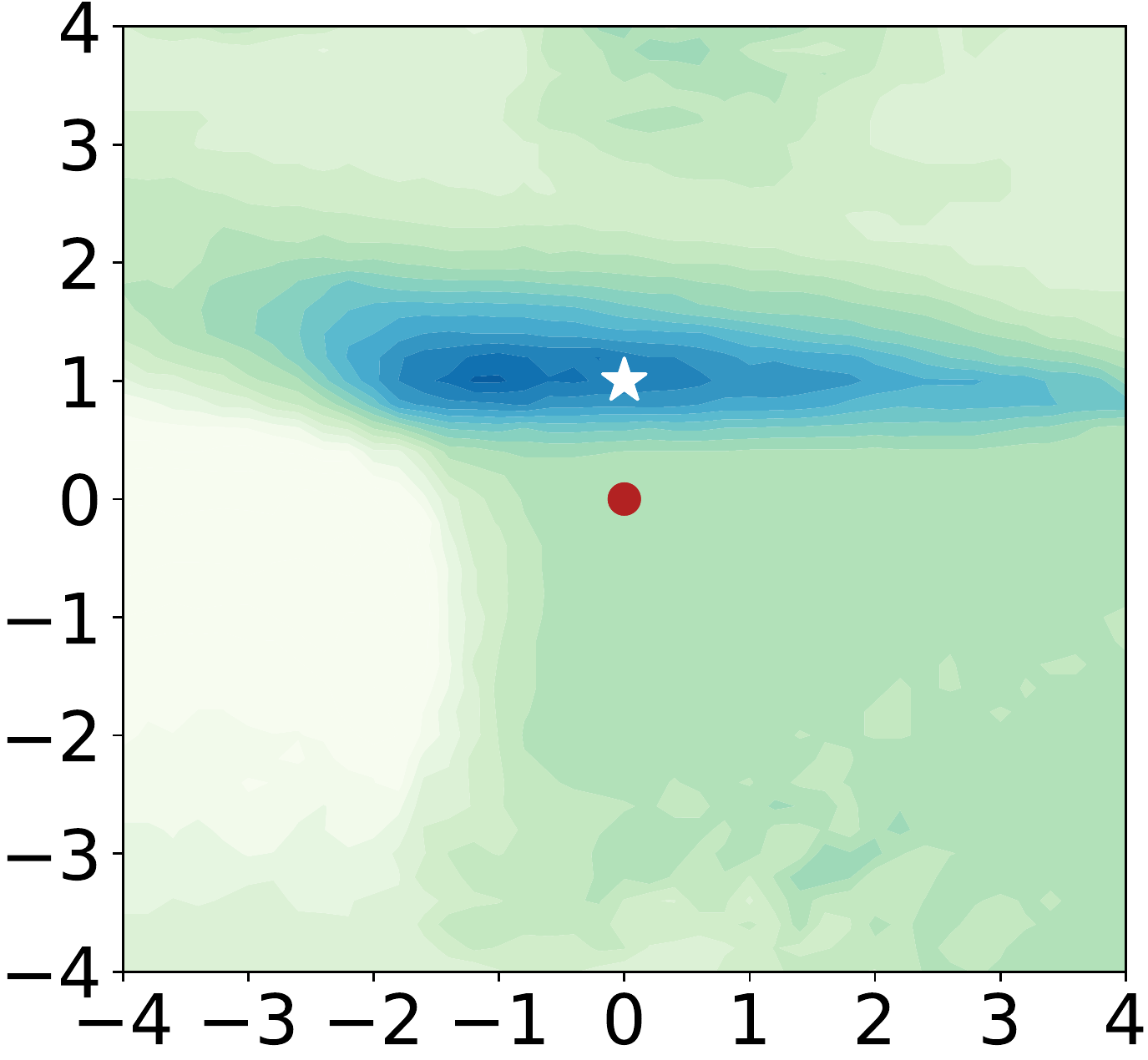}}&
\raisebox{-0.5\height}{\includegraphics[width=0.492\columnwidth]{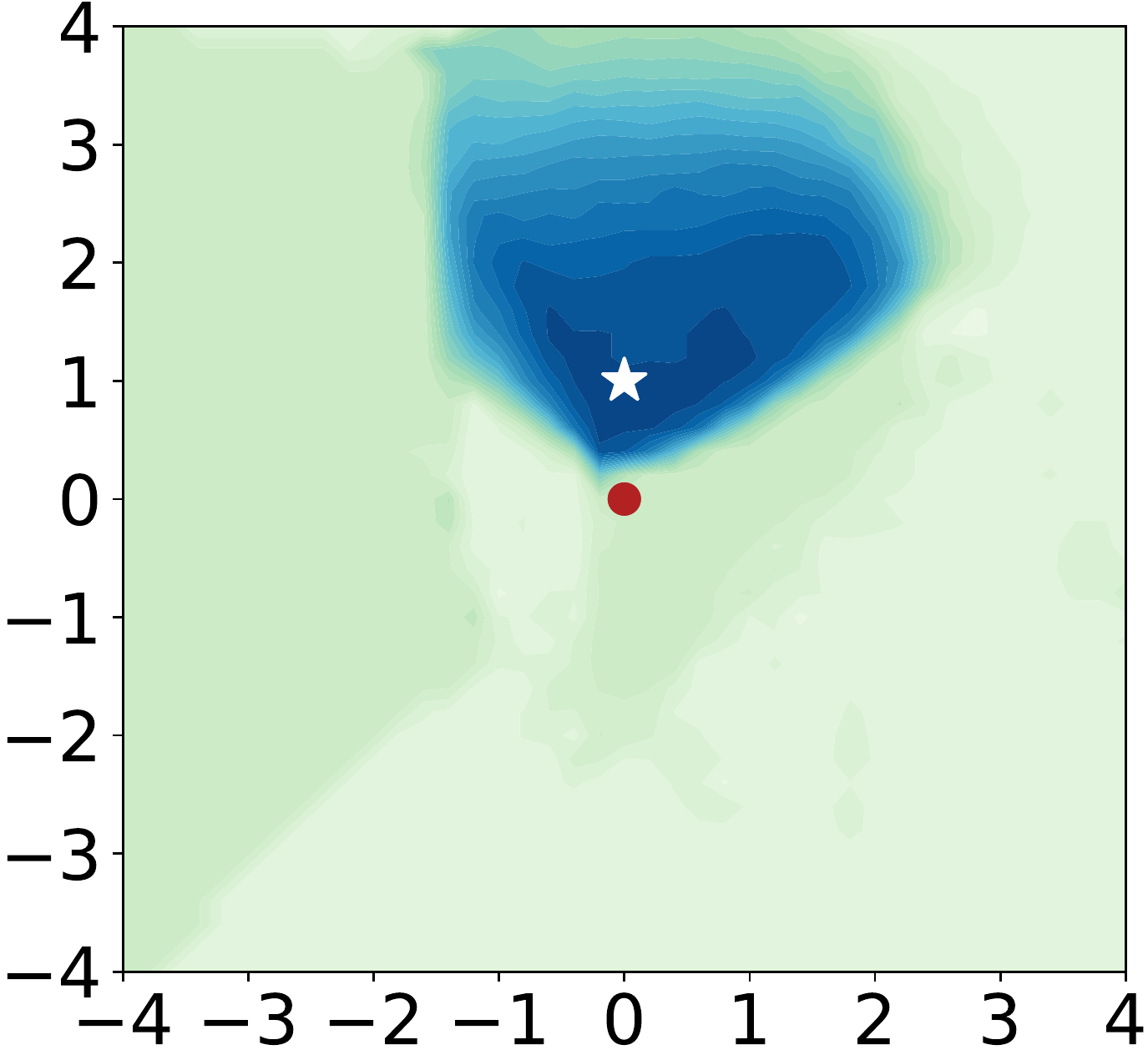}}&
\raisebox{-0.5\height}{\includegraphics[width=0.492\columnwidth]{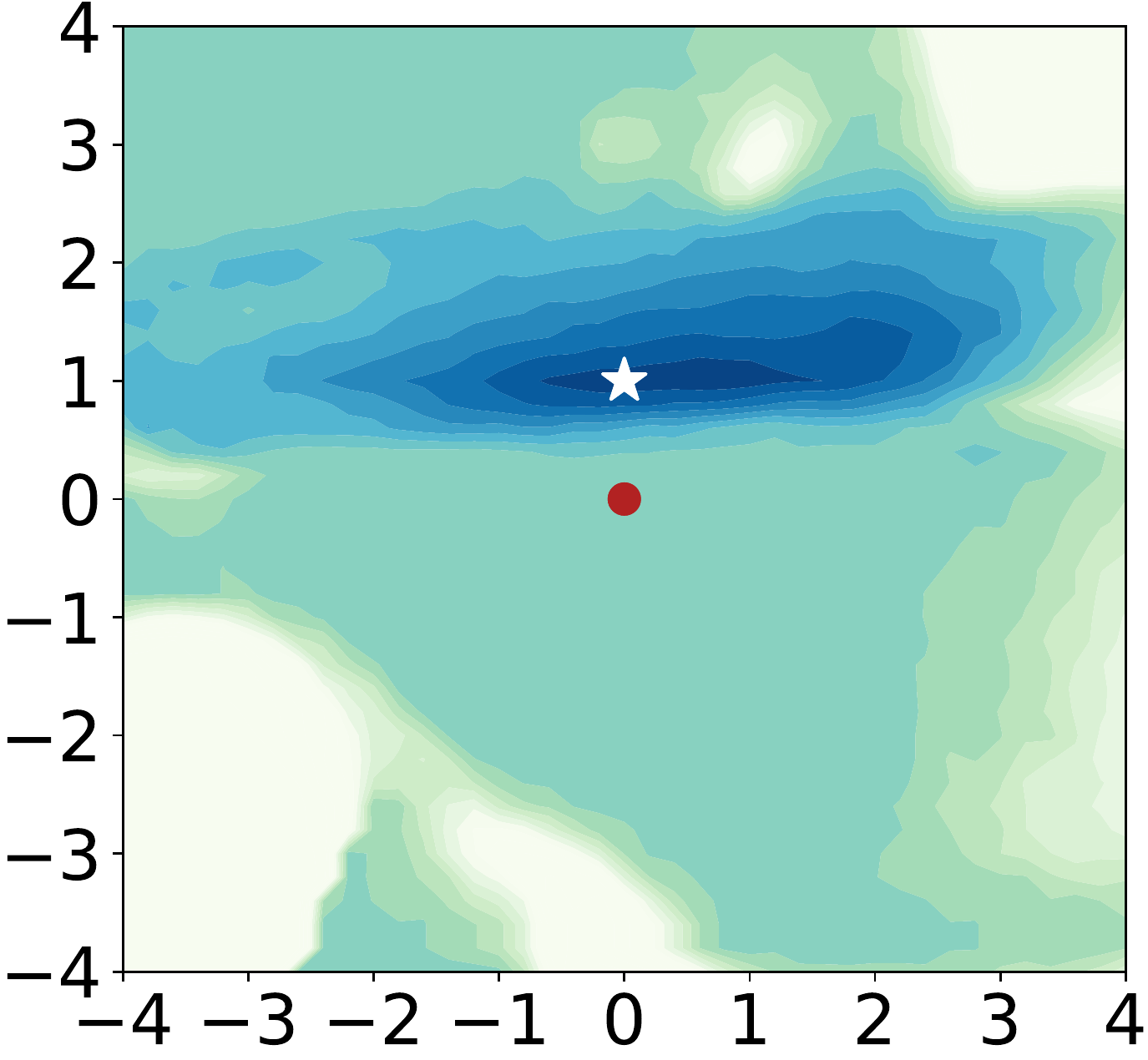}}\\
\addlinespace[2mm]
&
\includegraphics[width=0.3\columnwidth]{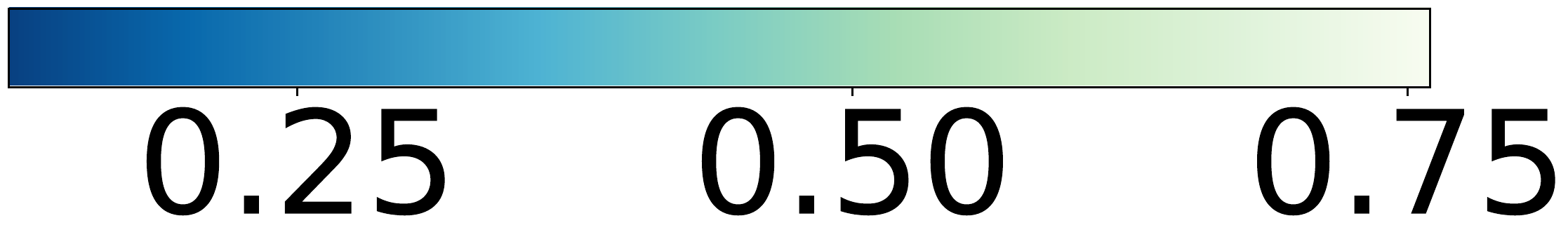}&
\includegraphics[width=0.3\columnwidth]{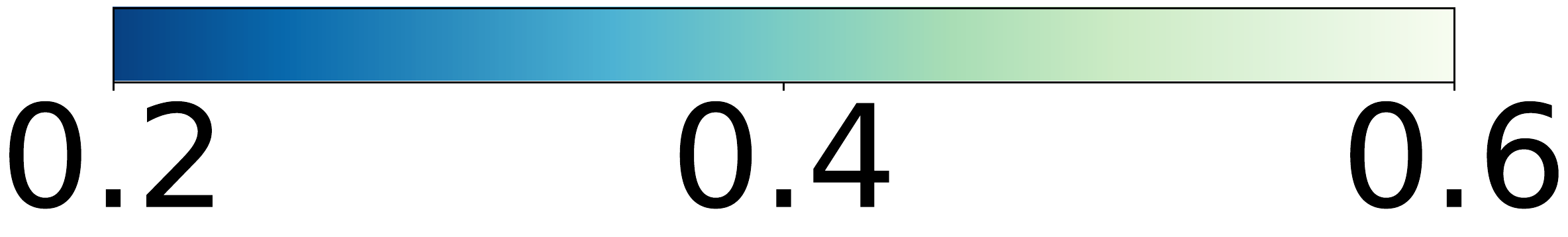}&
\includegraphics[width=0.3\columnwidth]{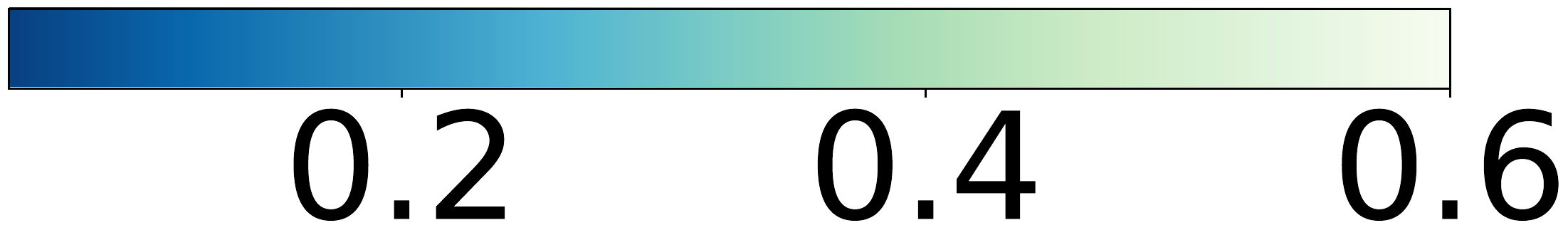}&
\includegraphics[width=0.3\columnwidth]{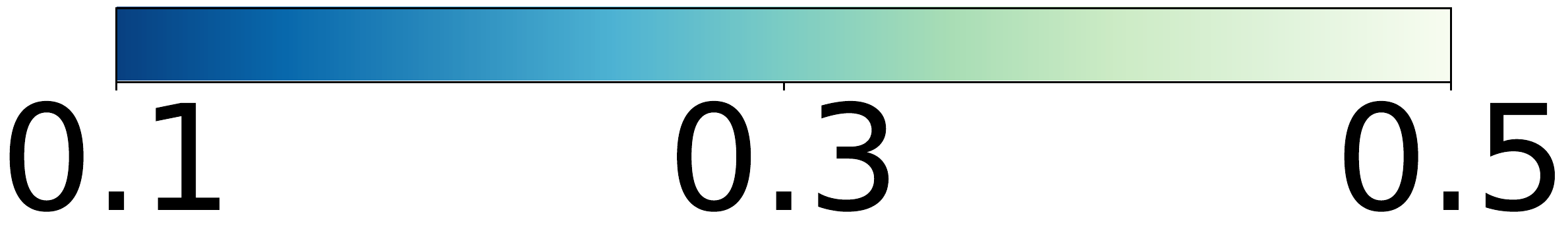}\\
\end{tabular}
\caption{Two-dimensional generalization error surfaces of training from scratch (top) and fine-tuning BERT (bottom).
The dot-shaped point and the star-shaped point indicate the generalization errors at the beginning and at the end of the training procedure, respectively.}
\label{fig:test_error_surface} 
\end{figure*}

To further understand the effects of geometry of loss functions to the generalization capability, we make comparisons between the training loss surfaces and the generalization error surfaces on different datasets.
The classification error rate on the development set is used as an indicator of the generalization capability.

As shown in Figure~\ref{fig:test_error_surface}, we find the end points of fine-tuning BERT fall into the wide areas with smaller generalization error.
The results show that the generalization error surfaces are consistent with the corresponding training loss surfaces on the datasets, i.e., smaller training loss tends to decrease the error on the development set. Moreover, the fine-tuned BERT models tend to stay approximately optimal under subtle perturbations.
The visualization results also indicate that it is preferred to converge to wider and more flat local optima, as the training loss surface and the generalization error surface are shifted with respect to each other~\cite{izmailov2018averaging}.
In contrast, training from scratch obtains thinner optimum areas and poorer generalization than fine-tuning BERT, especially on the datasets with relatively small data size (such as MRPC, and RTE). Intuitively, the thin and sharp optima on the training loss surfaces are hard to be migrated to the generalization surfaces.

For training from scratch, it is not surprising that on larger datasets (such as MNLI, and SST-2) the generalization error surfaces are more consistent with the training loss surfaces.
The results suggest that training the model from scratch usually requires more training examples to generalize better compared with fine-tuning BERT.

\section{Lower Layers of BERT are More Invariant and Transferable}

The BERT-large model has $24$ layers. Different layers could have learned different granularities or perspectives of language during the pre-training procedure.
For example, \citet{pipline} observe that most local syntactic phenomena are encoded in lower layers while higher layers capture more complex semantics.
They also show that most examples can be classified correctly in the first few layers.
From above, we conjecture that lower layers of BERT are more invariant and transferable across tasks.

\begin{figure*}[t]
\centering
\small
\begin{tabular}{l@{\hspace*{\lengthd}}c@{\hspace*{\lengthd}}c@{\hspace*{\lengthd}}c@{\hspace*{\lengthd}}c@{\hspace*{\lengthd}}c}
&Low layers (0-7)&Middle layers (8-15)&High layers (16-23)& \\
\rotatebox[origin=c]{90}{MNLI}&
\raisebox{-0.5\height}{\includegraphics[width=0.5\columnwidth]{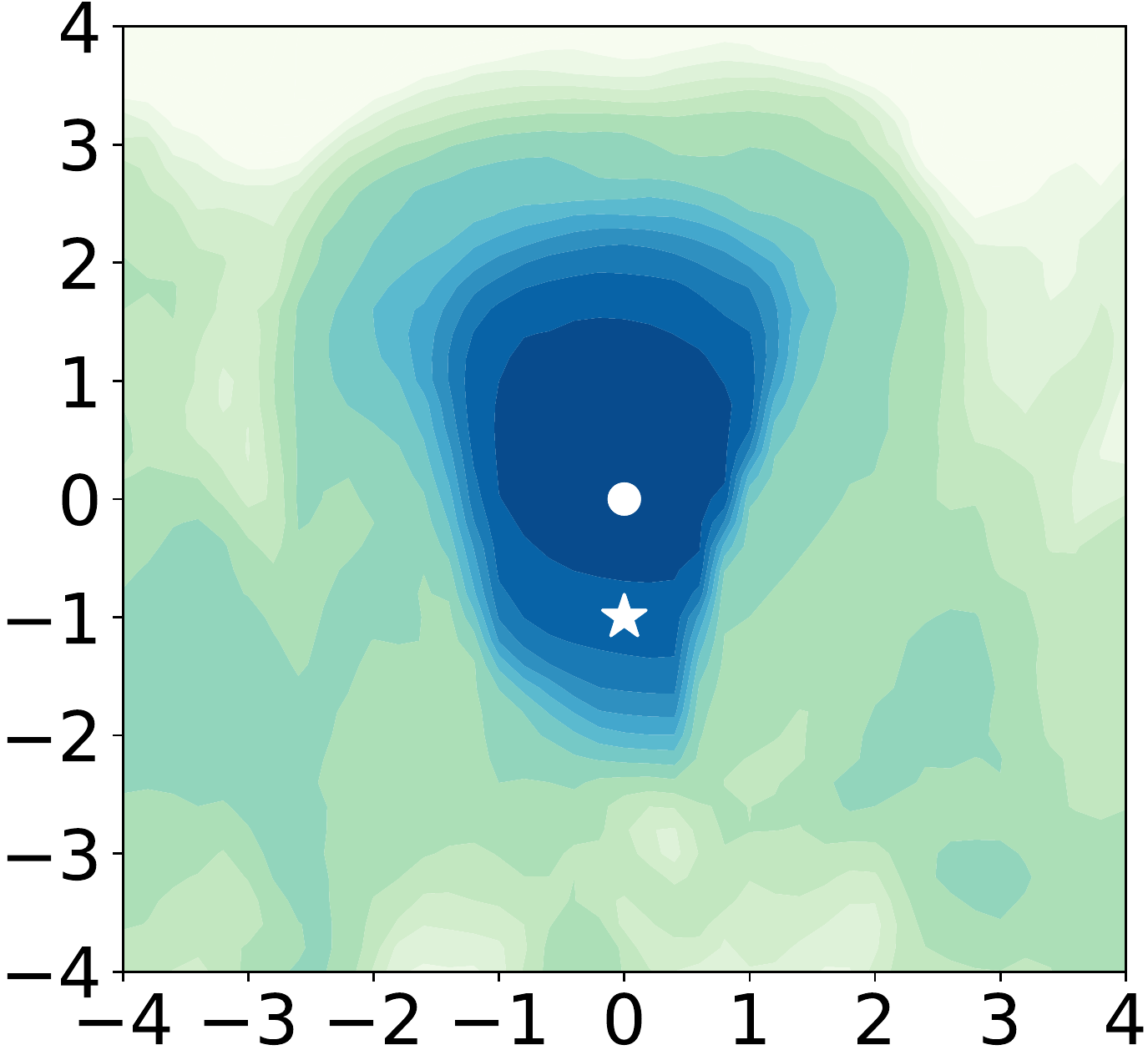}}&
\raisebox{-0.5\height}{\includegraphics[width=0.5\columnwidth]{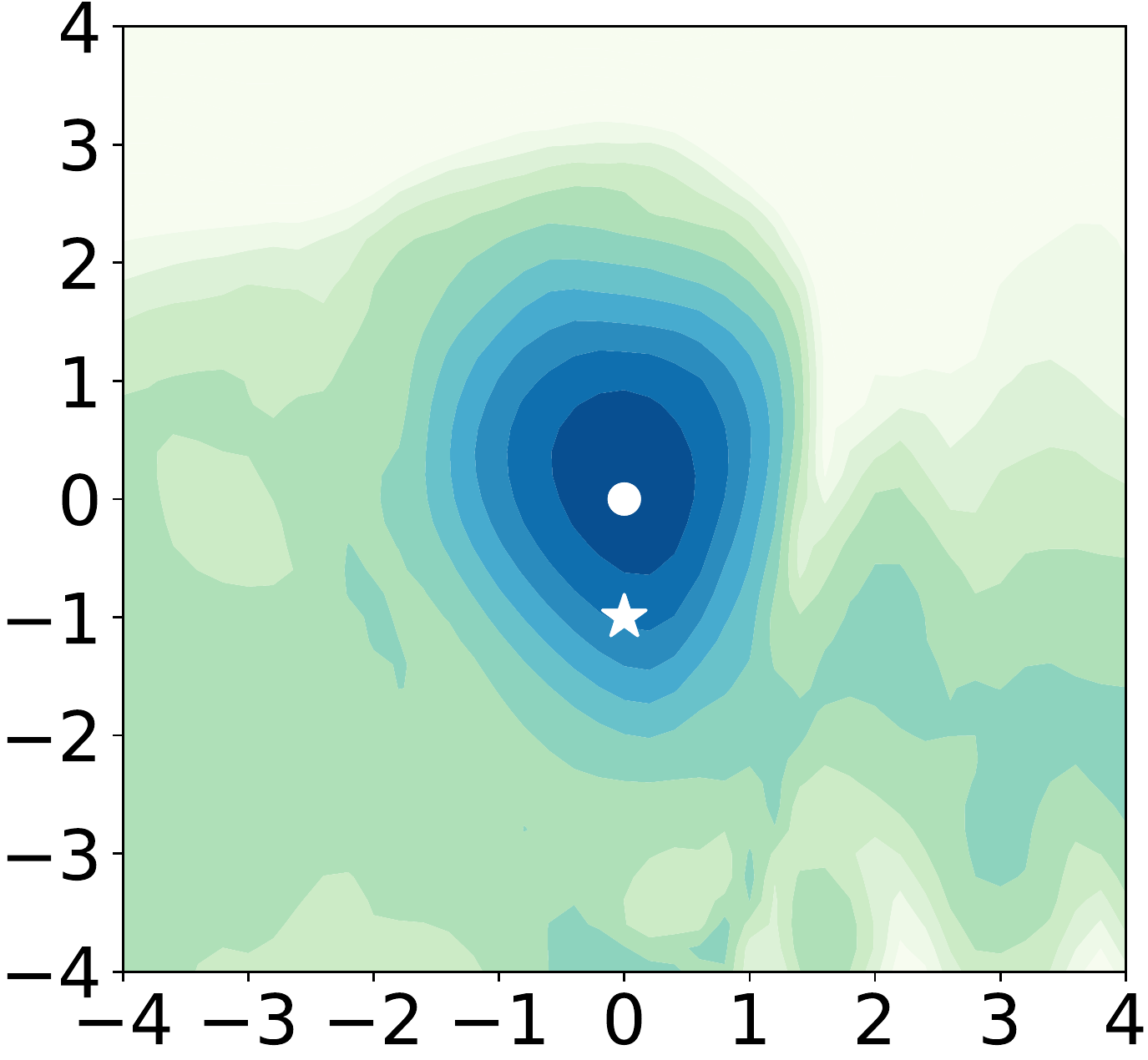}}&
\raisebox{-0.5\height}{\includegraphics[width=0.5\columnwidth]{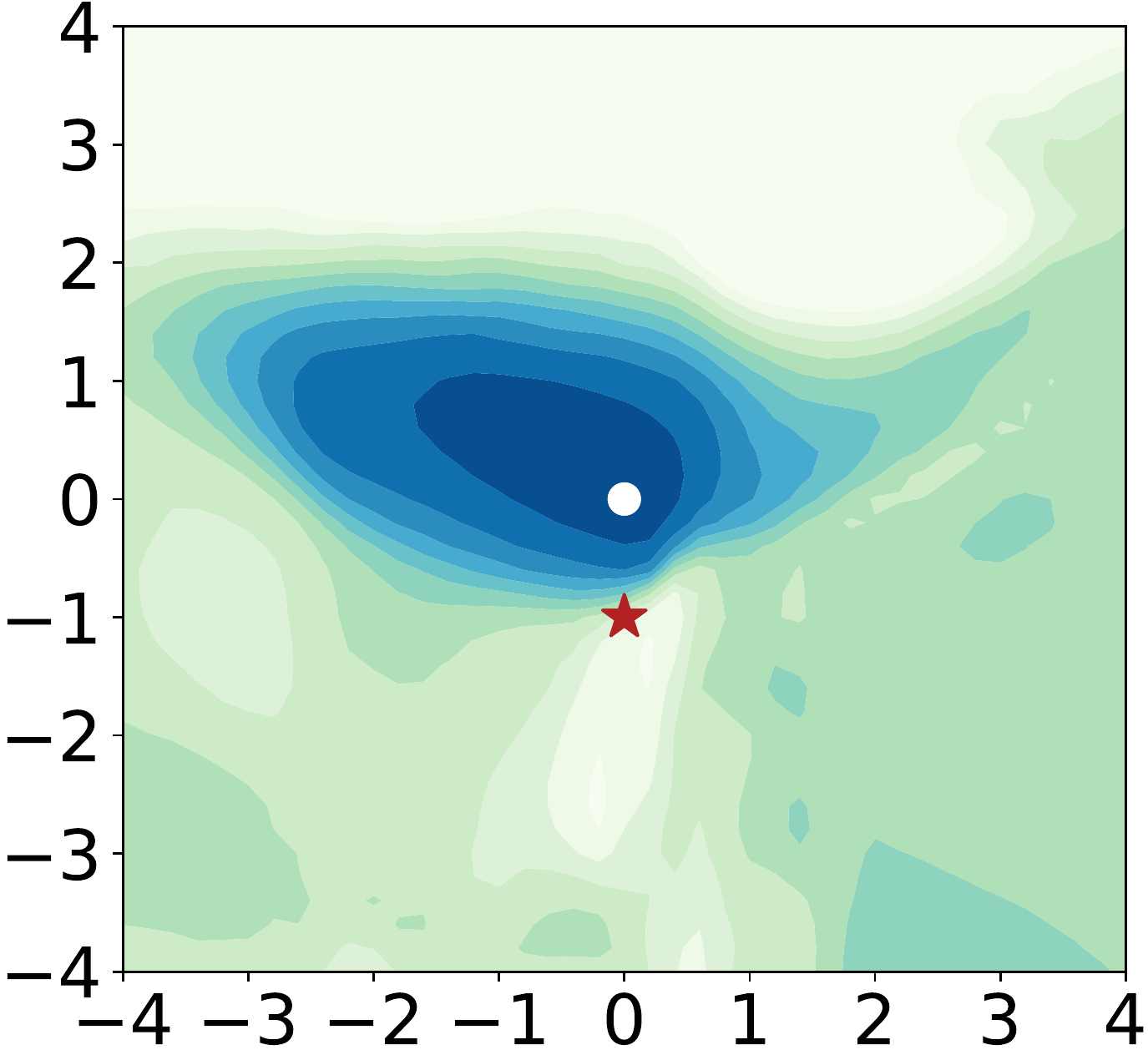}}&
\raisebox{-0.5\height}{\includegraphics[width=0.07\columnwidth]{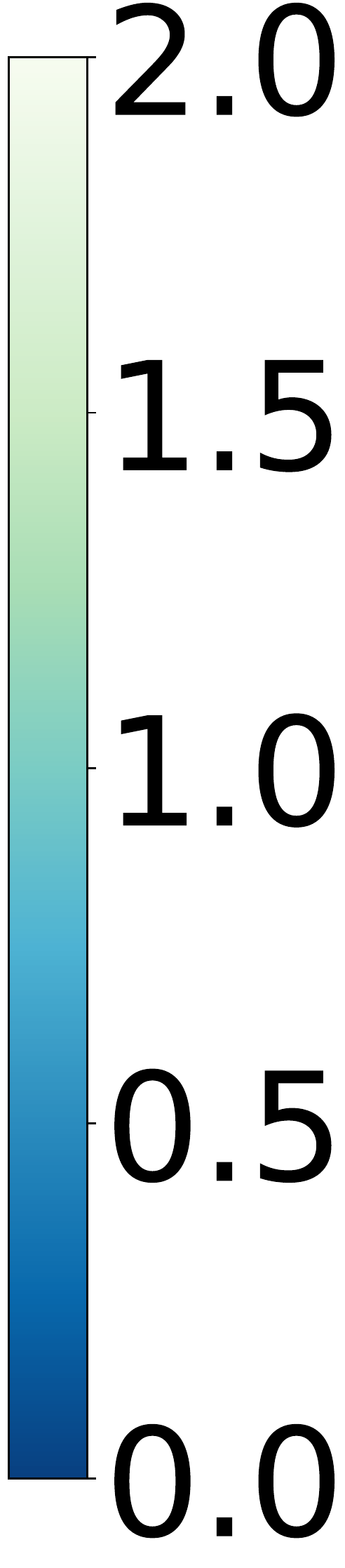}}\\
\addlinespace[0.3cm]
\rotatebox[origin=c]{90}{MRPC}&
\raisebox{-0.5\height}{\includegraphics[width=0.5\columnwidth]{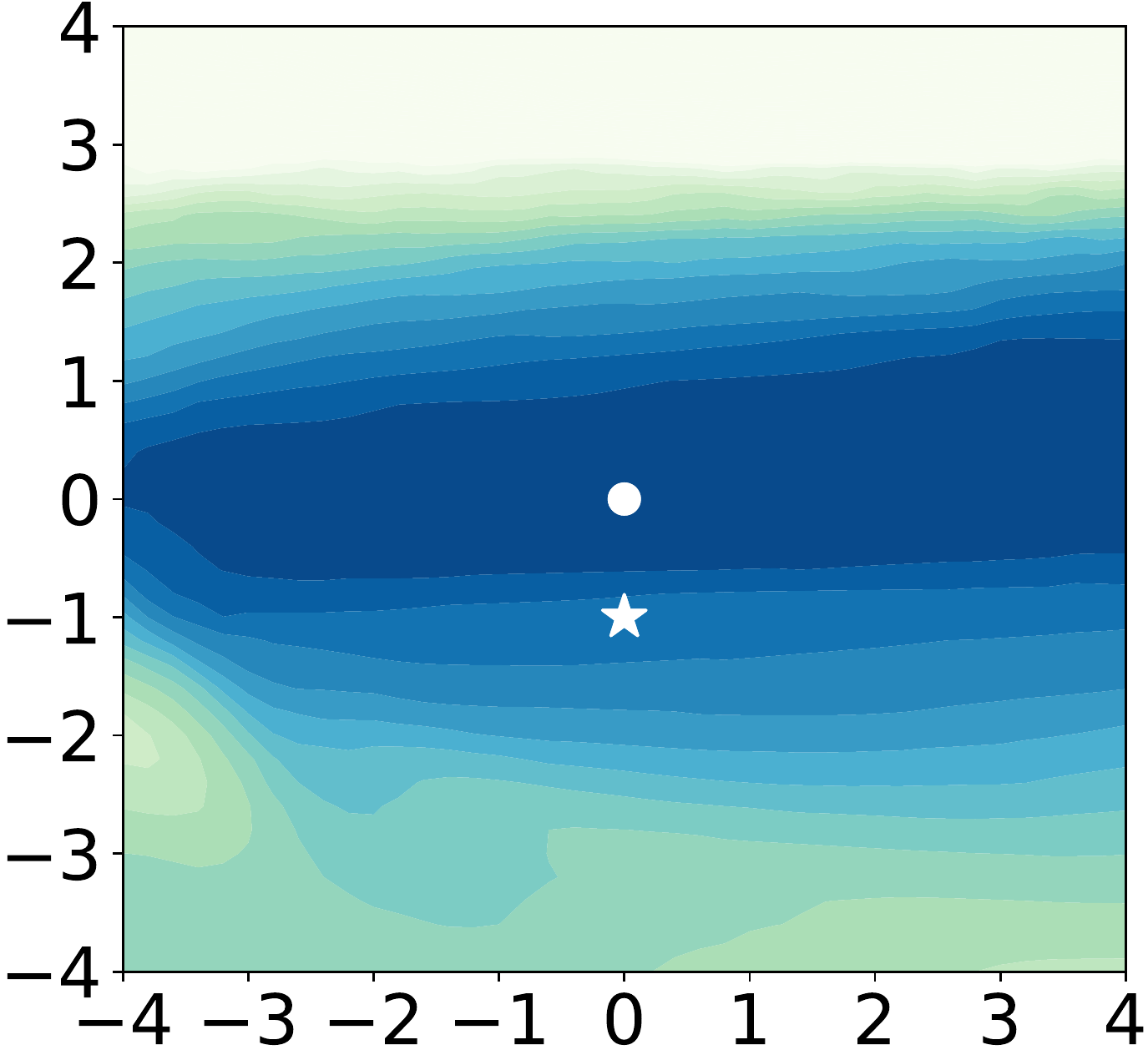}}&
\raisebox{-0.5\height}{\includegraphics[width=0.5\columnwidth]{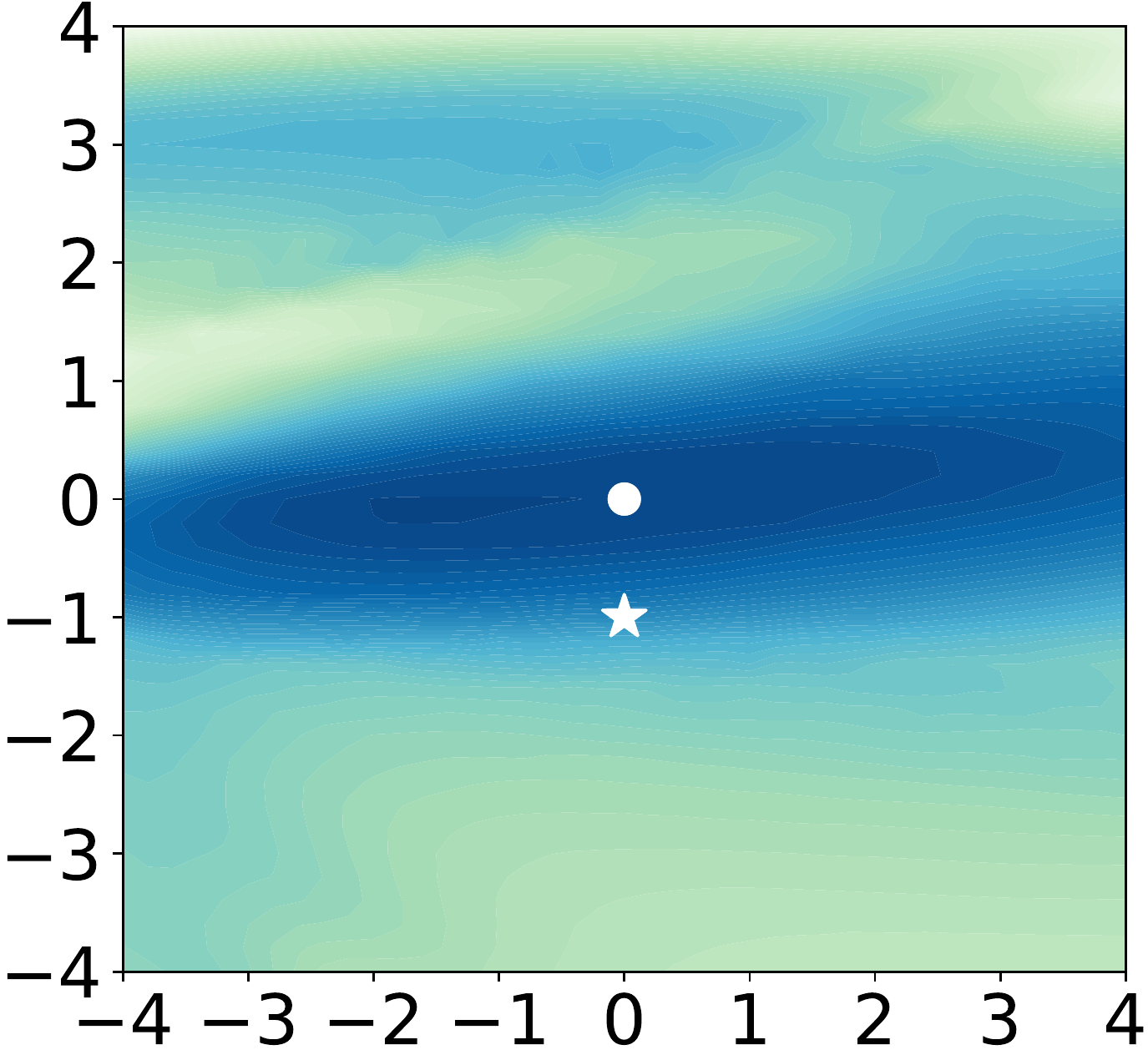}}&
\raisebox{-0.5\height}{\includegraphics[width=0.5\columnwidth]{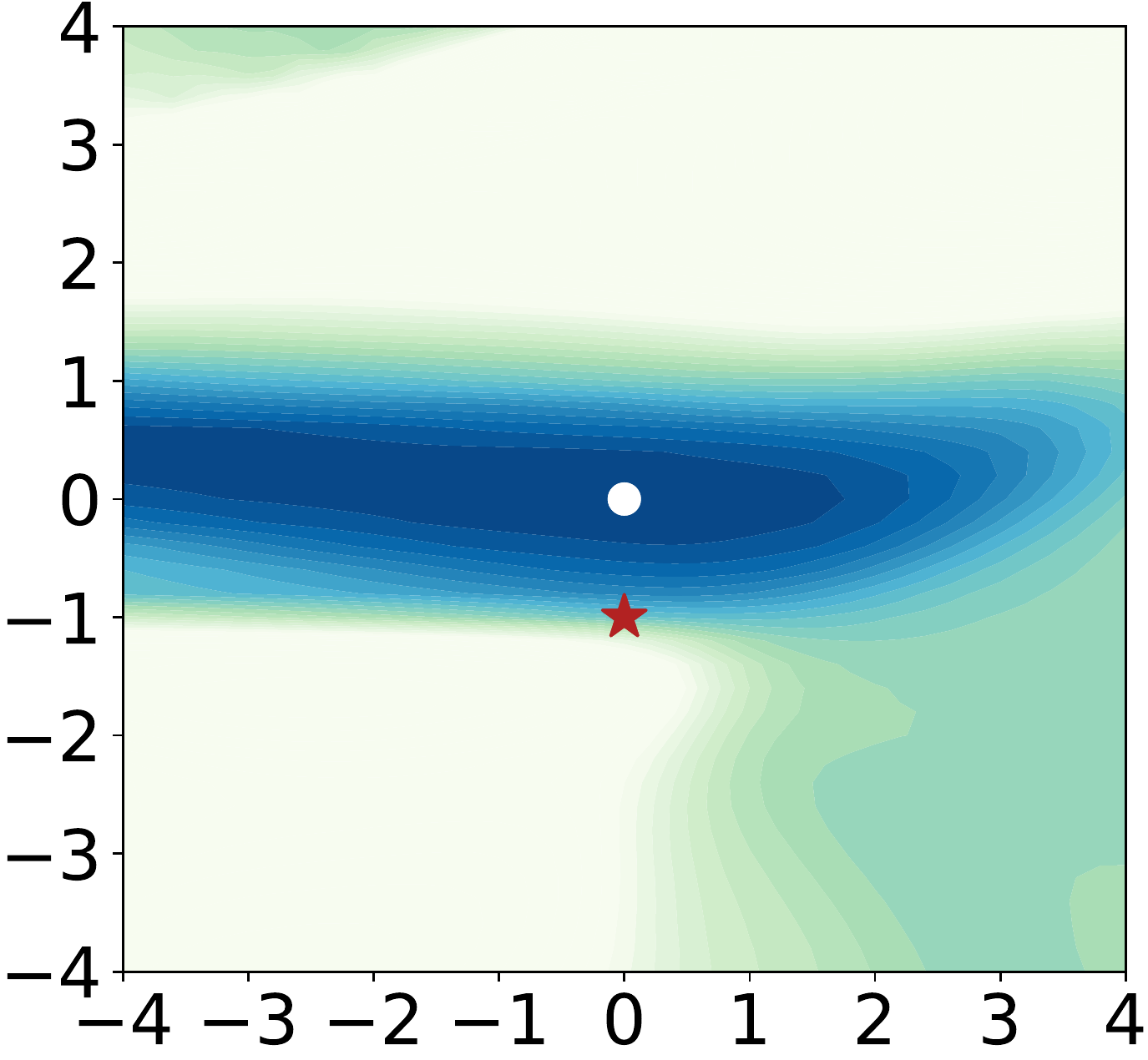}}&
\raisebox{-0.5\height}{\includegraphics[width=0.07\columnwidth]{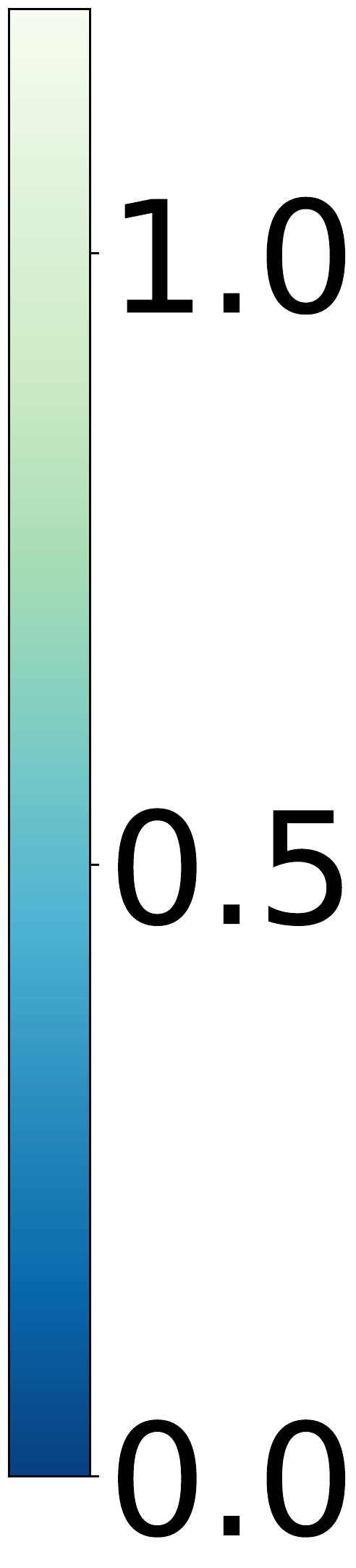}}\\
\end{tabular}
\caption{Layer-wise training loss surfaces on the MNLI dataset (top) and the MRPC dataset (bottom). The dot-shaped point represents the fine-tuned model. The star-shaped point represents the model with rollbacking different layer groups. Low layers correspond to the 0th-7th layers of BERT, middle layers correspond to the 8th-15th layers, and high layers correspond to the 16th-23rd layers.}
\label{fig:layer_surf_group} 
\end{figure*}

We divide the layers of the BERT-large model into three groups: low layers (0th-7th layer), middle layers (8th-15th layer), and high layers (16th-23rd layer).
As shown in Figure~\ref{fig:layer_surf_group}, we plot the two-dimensional loss surfaces with respect to different groups of layers (i.e., parameter subspace instead of all parameters) around the fine-tuned point.
To be specific, we modify the loss surface function in Section~\ref{subsc:loss_surface} to $f(\alpha, \beta)=\mathcal{J}(\bm{\theta}_1+\alpha \bm{\delta}_1^{G}+\beta \bm{\delta}_2^{G})$, where $\bm{\theta}_1$ represents the fine-tuned parameters, $G \in \{$low layers, middle layers, high layers$\}$, and the optimization direction of the layer group is used as the axis.
On the visualized loss landscapes, $f(0,0)$ corresponds to the loss value at the fine-tuned point.
Besides, $f(-1,0)$ corresponds to the loss value with the corresponding layer group rollbacked to its original values in the pre-trained BERT model.

Figure~\ref{fig:layer_surf_group} shows that the loss surface with respect to lower layers has the wider local optimum along the optimization direction.
The results demonstrate that rollbacking parameters of lower layers to their original values (the star-shaped points in Figure~\ref{fig:layer_surf_group}) does not dramatically hurt the model performance.
In contrast, rollbacking high layers makes the model fall into the region with high loss.
This phenomenon indicates that the optimization of high layers is critical to fine-tuning whereas lower layers are more invariant and transferable across tasks.

\begin{table}[t]
\centering
\begin{tabular}{lll@{\hspace*{\lengtha}}l@{\hspace*{\lengtha}}l}
\toprule
\multirow{2}{*}{Dataset} & \multirow{2}{*}{BERT} & \multicolumn{3}{c}{Layer Rollback} \\ \cmidrule{3-5}
                         &    & 0-7    & 8-15  & 16-23   \\ \midrule
MNLI                     & 86.54   & \tabincell{l}{86.73 \\ (+0.19)}  & \tabincell{l}{84.71 \\ (-1.83)}   & \tabincell{l}{32.85 \\ (-53.88)}  \\
RTE                      & 75.45   & \tabincell{l}{73.29 \\ (-2.16)}  & \tabincell{l}{70.04 \\ (-5.41)}   & \tabincell{l}{47.29 \\ (-28.16)}  \\
SST-2                    & 94.04   & \tabincell{l}{93.69 \\ (-0.35)}  & \tabincell{l}{93.12 \\ (-0.92)}   & \tabincell{l}{59.29 \\ (-34.75)}  \\
MRPC                     & 90.20   & \tabincell{l}{87.99 \\ (-2.21)}  & \tabincell{l}{80.15 \\ (-10.05)}   & \tabincell{l}{78.17 \\ (-12.03)}  \\
\bottomrule
\end{tabular}
\caption{Accuracy on the development sets. The second column represents the fine-tuned BERT models on the specific datasets. The last column represents the fine-tuned models with rollbacking different groups of layers.}
\label{tbl:rb_group}
\end{table}

In order to make a further verification, we rollback different layer groups of the fine-tuned model to the parameters of the original pre-trained BERT model.
The accuracy results on the development set are presented in Table \ref{tbl:rb_group}.
Similar to Figure~\ref{fig:layer_surf_group}, the generalization capability does not dramatically decrease after rollbacking low layers or middle layers.
Rollbacking low layers (0th-7th layer) even improves the generalization capability on the MNLI dataset.
By contrast, rollbacking high layers hurts the model performance.
Evaluation results suggest that low layers that are close to input learn more transferable representations of language, which makes them more invariant across tasks.
Moreover, high layers seem to play a more important role in learning task-specific information during fine-tuning.

\section{Related Work}

Pre-trained contextualized word representations learned from language modeling objectives, such as CoVe~\cite{mccann2017learned}, ELMo~\cite{elmo}, ULMFit~\cite{howard2018universal}, GPT~\cite{gpt,gpt2}, and BERT~\cite{bert}, have shown strong performance on a variety of natural language processing tasks.


Recent work of inspecting the effectiveness of the pre-trained models~\cite{syntax1,syntax2,probing1,probing2} focuses on analyzing the syntactic and semantic properties.
\citet{probing1} and~\citet{probing2} suggest that pre-training helps the models to encode much syntactic information and many transferable features through evaluating models on several probing tasks.
\citet{assessingBert} assesses the syntactic abilities of BERT and draws the similar conclusions.
Our work explores the effectiveness of pre-training from another angle.
We propose to visualize the loss landscapes and optimization trajectories of the BERT fine-tuning procedure. The visualization results help us to understand the benefits of pre-training in a more intuitive way.
More importantly, the geometry of loss landscapes partially explains why fine-tuning BERT can achieve better generalization capability than training from scratch.

\citet{probing2} find that different layers of BERT exhibit different transferability.
\citet{tuneornot} show that the classification tasks build up information mainly in the intermediate and last layers of BERT.
\citet{pipline} observe that low layers of BERT encode more local syntax, while high layers capture more complex semantics.
\citet{layerequal} also show that not all layers of a deep neural model have equal contributions to model performance.
We draw the similar conclusion by visualizing layer-wise loss surface of BERT on downstream tasks.
Besides, we find that low layers of BERT are more invariant and transferable across datasets.

In the computer vision community, many efforts have been made to visualize the loss function, and figure out how the geometry of a loss function affects the generalization~\cite{curve1d, surf2, visualloss}.
\citet{flatminima} define the flatness as the size of the connected region around a minimum.
\citet{sharpness} characterize the definition of flatness using eigenvalues of the Hessian, and conclude that small-batch training converges to flat minima, which leads to good generalization.
\citet{visualloss} propose a filter normalization method to reduce the influence of parameter scale, and show that the sharpness of a minimum correlates well with generalization capability.
The assumption is also used to design optimization algorithms~\cite{widevalley,izmailov2018averaging}, which aims at finding broader optima with better generalization than standard SGD.

\section{Conclusion}

We visualize the loss landscapes and optimization trajectories of the BERT fine-tuning procedure, which aims at inspecting the effectiveness of language model pre-training.
We find that pre-training leads to wider optima on the loss landscape, and eases optimization compared with training from scratch.
Moreover, we give evidence that the pre-training-then-fine-tuning paradigm is robust to overfitting.
We also demonstrate the consistency between the training loss surfaces and the generalization error surfaces, which explains why pre-training improves the generalization capability.
In addition, we find that low layers of the BERT model are more invariant and transferable across tasks.

All our experiments and conclusions were derived from BERT fine-tuning.
A further understanding of how multi-task training with BERT~\cite{mt-dnn} improves fine-tuning, and how it affects the geometry of loss surfaces are worthy of exploration, which we leave to future work.
Moreover, the results motivate us to develop fine-tuning algorithms that converge to wider and more flat optima, which would lead to better generalization on unseen data.
In addition, we would like to apply the proposed methods for other pre-trained models.

\section*{Acknowledgements}
The work was partially supported by National Natural Science Foundation of China (NSFC) [Grant No. 61421003].

\bibliography{finetuning}
\bibliographystyle{acl_natbib}

\end{document}